\documentclass[10pt,twocolumn,letterpaper]{article}

\usepackage{iccv}
\usepackage{times}
\usepackage{epsfig}
\usepackage{graphicx}
\usepackage{amsmath}
\usepackage{amssymb}
\usepackage{booktabs}
\usepackage{threeparttable}
\usepackage{multirow}
\usepackage{diagbox}
\usepackage{caption}
\usepackage{subfigure}
\usepackage{multicol}
\usepackage{rotating}

\usepackage[pagebackref=true,breaklinks=true,letterpaper=true,colorlinks,bookmarks=false]{hyperref}

\iccvfinalcopy 

\def\iccvPaperID{****} 

\ificcvfinal\fi

\begin{document}

\title{Structure-Preserving Deraining with Residue Channel Prior Guidance}

\author{Qiaosi Yi$^{1}$,\,Juncheng Li$^1$,\,Qinyan Dai$^1$,\,Faming Fang$^{1}\thanks{Corresponding author}$\,,\,Guixu Zhang$^1$,\,and Tieyong Zeng$^2$\\
$^{1}$Shanghai Key Laboratory of Multidimensional Information Processing,\\ and School of Computer Science and Technology, East China Normal University, Shanghai, China\\
$^{2}$Department of Mathematics, The Chinese University of Hong Kong, Hong Kong, China\\
{\tt\small qiaosiyijoyies@gmail.com, cvjunchengli@gmail.com, 649310204@qq.com}\\
\vspace{-10pt}
{\tt\small fmfang@cs.ecnu.edu.cn, gxzhang@cs.ecnu.edu.cn, zeng@math.cuhk.edu.hk}
}
\maketitle
\ificcvfinal\thispagestyle{empty}\fi

\begin{abstract}
   Single image deraining is important for many high-level computer vision tasks since the rain streaks can severely degrade the visibility of images, thereby affecting the recognition and analysis of the image. Recently, many CNN-based methods have been proposed for rain removal. Although these methods can remove part of the rain streaks, it is difficult for them to adapt to real-world scenarios and restore high-quality rain-free images with clear and accurate structures. To solve this problem, we propose a Structure-Preserving Deraining Network (SPDNet) with RCP guidance. SPDNet directly generates high-quality rain-free images with clear and accurate structures under the guidance of RCP but does not rely on any rain-generating assumptions. Specifically, we found that the RCP of images contains more accurate structural information than rainy images. Therefore, we introduced it to our deraining network to protect structure information of the rain-free image. Meanwhile, a Wavelet-based Multi-Level Module (WMLM) is proposed as the backbone for learning the background information of rainy images and an Interactive Fusion Module (IFM) is designed to make full use of RCP information. In addition, an iterative guidance strategy is proposed to gradually improve the accuracy of RCP, refining the result in a progressive path. Extensive experimental results on both synthetic and real-world datasets demonstrate that the proposed model achieves new state-of-the-art results. Code: https://github.com/Joyies/SPDNet
\end{abstract}

\vspace{-20pt}
\section{Introduction}\label{introduce}

\begin{figure}[t]
	\centering
    \includegraphics[scale=0.21]{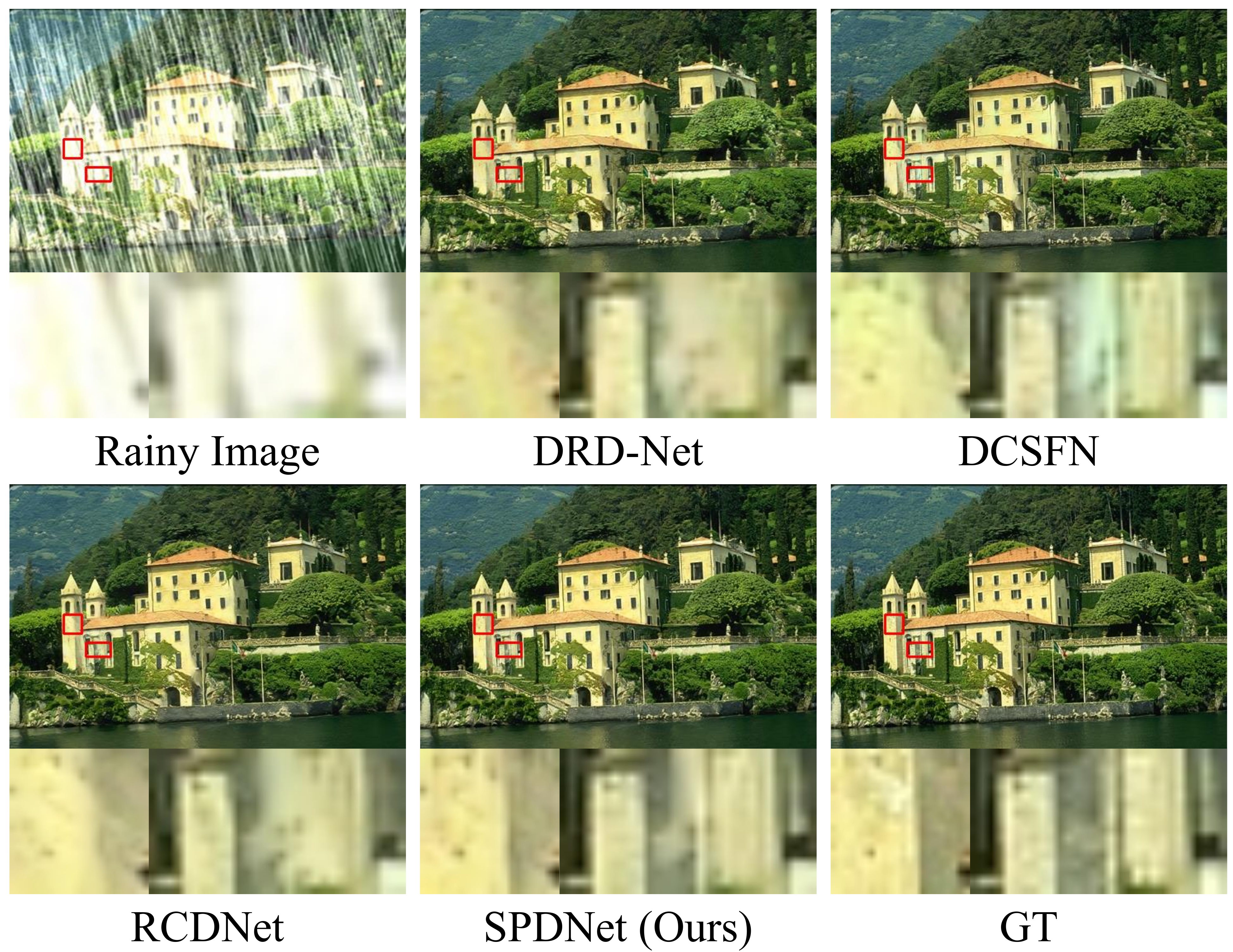}
    \setlength{\abovecaptionskip}{-3pt}
    \vspace{-3pt}
	\caption{An example of image deraining results. Obviously, SPDNet can reconstruct high-quality image with clear and accurate structure.}
	\label{fig:head}
	\vspace{-14pt}
\end{figure}

Single image deraining (SID) aims to reconstruct a visually pleasing image from its corresponding rain-streaks-degraded image. In the past years, various methods have been proposed for SID. 
For traditional methods, many researchers~\cite{garg2005does, barnum2010analysis, bossu2011rain, chen2013generalized, xu2012improved, li2016rain, luo2015removing, GuMZZ17} focused on exploring the physical properties of the rain and background layers. Meanwhile, various priors have been proposed to regularize and separate them, such as layer priors with Gaussian mixture model (GMM)~\cite{li2016rain}, discriminative sparse coding (DSC)~\cite{luo2015removing}, and joint convolutional analysis and synthesis sparse representation (JCAS)~\cite{GuMZZ17}. Although these methods can effectively remove the rain streaks, they require complex iterative optimization to find the best solution.

Recently, convolutional neural networks (CNN) have achieved significant success in many computer vision tasks~\cite{simonyan2014very, he2016deep, li2019lightweight,peng2020cumulative}. Moreover, many CNN-based methods~\cite{FuHDLP17,fu2017removing, yang2017deep, li2018recurrent, zhang2018density, hu2019depth, wang2019spatial, zhang2019image, ren2019progressive, yang2019single, deng2020detail, wang2020model, jiang2020multi} have been proposed for rain removal, such as DDN~\cite{fu2017removing},  RESCAN~\cite{li2018recurrent}, PReNet~\cite{ren2019progressive}, DRDNet~\cite{deng2020detail}, and  RCDNet~\cite{wang2020model}. Albeit these methods have brought great performance improvements, they are difficult to remove all rain streaks and recover the structural information of images in complex scenarios (Fig.~\ref{fig:head}). This is because: (1). Most of them directly learn the mapping between the rainy image and the rain streaks layers. In other words, most methods predict rain streaks via the built CNN model and then subtract rain streaks from rainy images to get the final output. However, the density of rain streaks varies, which leads to excessive or insufficient removal of rain streaks, resulting in incomplete structural information of the reconstructed images. (2) These methods focus on learning the structure of rain streaks, but they pay less attention to learning the structure of objects and ignore the importance of image prior. 

\begin{figure}[t]
	\centering
	\begin{minipage}[c]{0.23\textwidth}
		\includegraphics[scale=0.126]{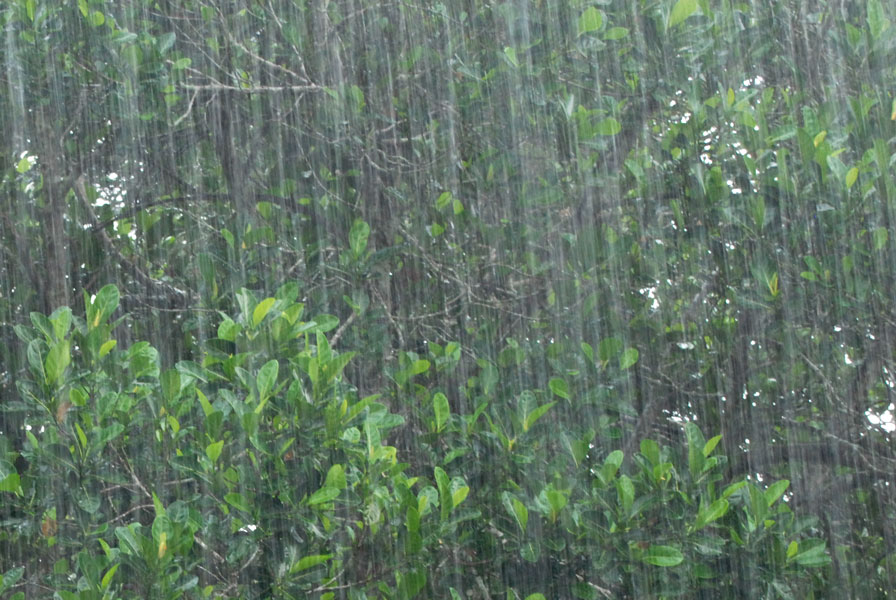}
		\centerline{A. Rainy Image}
	\end{minipage}
	\begin{minipage}[c]{0.23\textwidth}
		\includegraphics[scale=0.126]{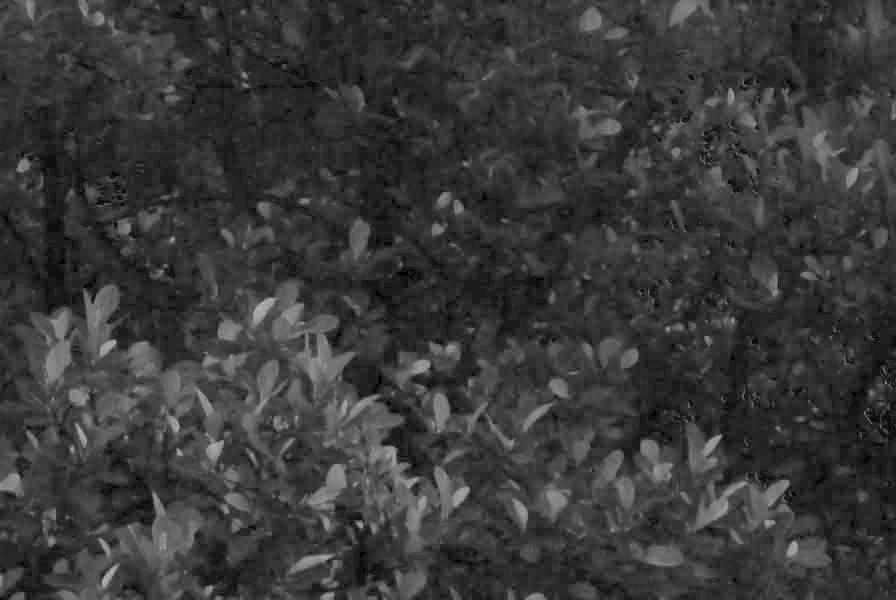}
		\centerline{B. RCP of A}
	\end{minipage}
    \vspace{-3pt}
	\caption{B is the residue channel prior (RCP) extracted from the rainy image. Obviously, even the RCP is extracted from the rainy image, it still contains clear structures.}
	\label{fig:rcp}
	\vspace{-14pt}
\end{figure}

To address the aforementioned issues, we aim to explore an image prior that can protect the structure of the image, and introduce the prior into the model to guide high-quality image reconstruction. According to our investigation, we found that the Total Variation (TV) prior will smooth texture details in the restored images, the sparse prior is usually difficult to model because it requires other domain knowledge, and the edge prior is difficult to obtain from the rainy image since the off-the-shelf edge detectors are sensitive to the rain streaks. In contrast, residue channel prior (RCP~\cite{li2018robust,li2019rainflow,li2020all}) show clear structures even extracted from the rainy image (Fig.~\ref{fig:rcp}). Moreover, compared with layer prior~\cite{luo2015removing} and ~\cite{hu2019depth}, RCP is the residual result of the maximum channel value and minimum channel value of the rainy image, calculated without any additional parameters. Hence, we adopt RCP to the image deraining task and propose a Structure-Preserving Deraining Network (SPDNet) with residue channel prior (RCP) guidance. SPDNet pays more attention to learning the background information and directly generates high-quality rain-free images with clear and accurate structures. To achieve this, an RCP guidance network and an iterative guidance strategy are proposed for structure-preserving deraining. Specifically, we design a wavelet-based multi-level module (WMLM) as the backbone of SPDNet to fully learn the background information in the rainy image. Meanwhile, an RCP extraction module and an interactive fusion module (IFM) are designed for RCP extraction and guidance, respectively. This is also the most important step in the SPDNet, which can highlight the structure of objects in the rainy images and promote generating high-quality rain-free images. In addition, we found that the RCP improves as image quality improves. Therefore, we propose a progressive reconstruction method, which means that the model extracts more accurate RCPs in the intermediate reconstruction stage to further guide image reconstruction. Under the iterative guidance of RCP, SPDNet can reconstruct high-quality rain-free images.

The main contributions of this paper are as follows:
\begin{itemize}
	\item We explore the importance of residue channel prior (RCP) for rain removal and propose a Structure-Preserving Deraining Network (SPDNet) with RCP guidance. Extensive experimental results show that SPDNet achieves new state-of-the-art results. 
	
	\item We propose an RCP extraction module and an Interactive Fusion Module (IFM) for RCP extraction and guidance, respectively. Meanwhile, an iterative guidance strategy is designed for progressive image reconstruction.
	
	\item We design a Wavelet-based Multi-Level Module (WMLM) as the backbone of SPDNet to learn the background of the area covered by the rain streak.
\end{itemize}

\begin{figure*}[http]
	\begin{center}
		\includegraphics[scale=0.3]{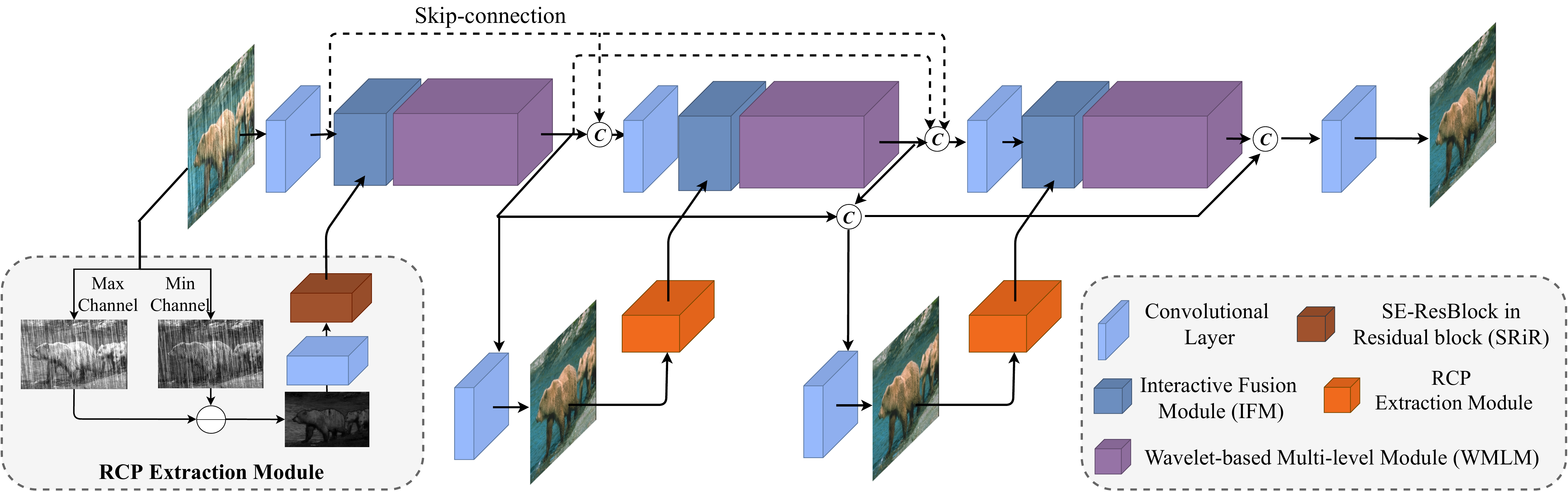}
	\end{center}
	\vspace{-15pt}
	\caption{The overall architecture of the proposed Structure-Preserving Deraining Network (SPDNet).}
	\label{fig:model}
	\vspace{-12pt}
\end{figure*}

\vspace{-10pt}
\section{Related Work}
In the past few years, many excellent rain removal methods have been proposed, which greatly promoted the development of image deraining. This part mainly focuses on several classical single image rain removal methods. A more detailed introduction can be found in~\cite{yang2020Single, li2021a, wang2020single}.

\subsection{Traditional Methods}
Traditional methods mainly use manually extracted features and priors to describe the features of rain streaks. For example, Kang~\etal~\cite{kang2011automatic} proposed a method that can decompose an image into low- and high-frequency parts by using a bilateral filter. Luo~\etal~\cite{luo2015removing} presented a discriminative sparse coding for separating rain streaks from the rainy image. Li~\etal~\cite{li2016rain} uses Gaussian mixture models(GMM) as the prior to separate the rain streaks. Wang~\etal~\cite{wang2017a} combines image decomposition and dictionary learning to remove rain or snow from images. Zhu~\etal~\cite{zhu2017joint} detects rain-dominant regions. The detected regions are utilized as a guidance image to help separate rain streaks from the background layer.

\subsection{CNN-based Methods}
Recently, many CNN-based image deraining networks have been proposed~\cite{yasarla2019uncertainty, wei2019semi,wang2019erl,yang2019scale,yang2020Single,yasarla2020syn2real,jiang2020multi,qian2018attentive, wang2020joint,fu2021rain}, and they have greatly promoted the development of this field. For example, Fu~\etal~\cite{FuHDLP17} proposed the firstly CNN-based model to remove rain streaks. Later, they~\cite{fu2017removing} also proposed a deeper network based on a deep residual network and use the image domain knowledge to remove rain streaks. Yang~\etal~\cite{yang2017deep} proposed a recurrent deep network for joint rain detection and removal to progressively remove rain streaks. Li~\etal~\cite{li2018recurrent} proposed a recurrent squeeze-and-excitation context aggregation network to make full use of contextual information. Deng~\etal~\cite{deng2020detail} introduced a detail-recovery image deraining network, which is composed of a rain residual network and a detail repair network. Jiang~\etal~\cite{jiang2020multi} designed a coarse-to-fine progressive fusion network and Wang~\etal~\cite{wang2020dcsfn} proposed a cross-scale fusion network based on inner-scale connection block. Moreover, Ren~\etal~\cite{ren2019progressive} proposed a progressive recurrent network (PReNet) by repeatedly unfolding a shallow ResNet. Wang~\etal~\cite{wang2020model} introduced a rain convolutional dictionary network, which continuously optimizes the model to obtain better rain streaks and rain-free images.

Although these methods can remove part of the rain streaks, they still cannot effectively remove all the rain streaks in complex situations. Meanwhile, these methods can hardly protect the structural information of the image thus cannot reconstruct high-quality rain-free images. Therefore, a method that can effectively remove rain streaks and protect the structure of the object is essential.

\vspace{10pt}

\section{Structure-Preserving Deraining Network}
In this paper, we propose a Structure-Preserving Deraining Network (SPDNet). As shown in Fig.~\ref{fig:model}, SPDNet uses the wavelet-based feature extraction backbone as the main structure and introduces a residue channel prior (RCP) guided mechanism for structure-preserving deraining. In SPDNet, the wavelet-based feature extraction backbone is designed for background information learning, which consists of a series of wavelet-based multi-level modules (WMLM). However, it is difficult to maintain clear structures of the reconstructed image by only using the wavelet-based feature extraction backbone. Hence, RCP is introduced to the model to provide additional auxiliary information, so that the structural information of the reconstructed image is protected. More details will be introduced in the following sections.
 
\begin{figure}
	\begin{center}
		\includegraphics[scale=0.4]{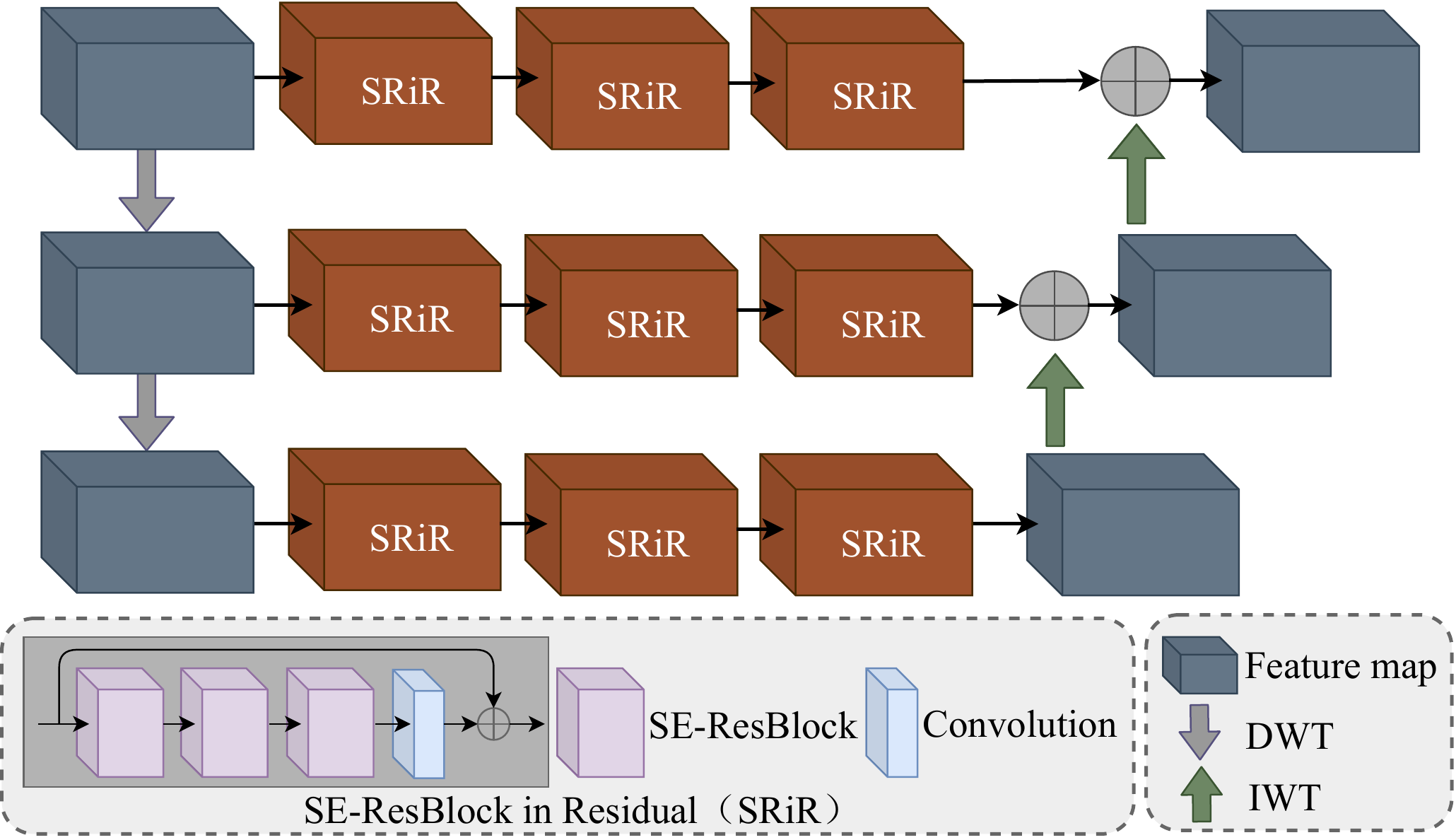}
	\end{center}
	\vspace{-10pt}
	\caption{The architecture of the proposed Wavelet-based Multi-Level Module (WMLM).}
	\label{fig:wmlm}
	\vspace{-16pt}
\end{figure}

\subsection{Wavelet-based Feature Extraction Backbone}
Due to the arbitrary size and density of the rain streaks,
the occluded region and the degree of occlusion of the object in the rainy image are unknown. To solve this problem, many methods~\cite{jiang2020multi, wang2020dcsfn, yi2021efficient} adopt the multi-level strategy to fully learning the features under different scales. Following these methods, we also adopt the multi-level strategy in our model. However, directly using the downsampling operation or deconvolution operation will cause a lot of information to be lost. Therefore, we propose a wavelet-based multi-level module (WMLM), which adopts the discrete wavelet transform (DWT) and Inverse DWT (IWT) in place of the simple downsampling and deconvolution operations. Moreover, DWT can capture both frequency and location information of feature maps~\cite{liu2018multi, daubechies1990wavelet, liu2020wavelet}, which may be helpful in preserving detailed textures. The architecture of WMLM is shown in Fig.~\ref{fig:wmlm}. 

In WMLM, we firstly use DWT to obtain multiple rainy image features with different scales, and adopt a convolution to resize the feature channel ($\widetilde{\mathcal{F}}_0, \widetilde{\mathcal{F}}_1, \widetilde{\mathcal{F}}_2$) as: 
\begin{equation}
    \left\{\begin{array}{l}
\widetilde{\mathcal{F}}_i = \widetilde{\mathcal{F}}, \quad \text{if} \,\,\, i=0,\\\\
\widetilde{\mathcal{F}}_i = Conv(DWT(\widetilde{\mathcal{F}}_{i-1})), \quad \text{if} \,\,\, i\textgreater0,\\
\end{array}\right.
\end{equation}
where $\widetilde{\mathcal{F}}$ denotes the input of WMLM. Next, we adopt three SE-ResBlock in Residual block (SRiR)~\cite{zhang2018image} to learn the background information of the rainy image. As shown in Fig.~\ref{fig:wmlm}, SRiR is similar to the ResBlock~\cite{he2016deep}, which uses the SE-Resblock~\cite{hu2018squeeze} to replace the convolution in ResBlock. 
\begin{equation}
    \widetilde{\mathcal{F}}_i^s = SRiR(\widetilde{\mathcal{F}_i}), \quad i=0,1,2.
\end{equation}
Lastly, we use a convolution to resize the smaller size features channel and up-sampled it by IWT. Meanwhile, they are added to the output of the previous level:
\begin{equation}
    \widetilde{\mathcal{F}}_{i-1}^s = IWT(Conv(\widetilde{\mathcal{F}}_i^s)) + \widetilde{\mathcal{F}}_{i-1}^s, \quad i = 2,1,
\end{equation}

With the help of WMLM, the model can effectively learn the background information of rainy images and reconstruct relatively clear images.

\subsection{RCP Guided Structure-Preserving Deraining}
Although the basic model can reconstruct relatively clear rain-free images, it is found that the object structure in the reconstructed image has also been damaged. In order to solve this problem, we recommend introducing additional image priors to protect the object structure. Therefore, we suggest introducing residue channel prior (RCP) to the model to achieve structure-preserving deraining. In addition, an Interactive Fusion Module (IFM) and an iterative guidance strategy are proposed to make full use of RCP information, so that high-quality rain-free images can be reconstructed.

\subsubsection{Residue Channel Prior}
As shown in Fig.~\ref{fig:rcp}, RCP is free of rain streaks and contains only a transformed version of the background details, which is the residual result of the maximum channel value and minimum channel value of the rainy image.

According to Li~\etal~\cite{li2018robust}, the colored-image intensity of a rain streak image can be expressed as:
\begin{equation}
\tilde{\mathcal{O}}({x})=t \beta_{r s}({x}) \mathcal{B} \alpha+(T-t) \mathcal{R} \boldsymbol{\pi},
\label{eq:i}
\end{equation}
where $\tilde{\mathcal{O}}({x})$ is the color vector representing the colored intensity. $\beta_{rs}$ is composed of refraction, specular reflection, and internal reflection coefficients of raindrops~\cite{garg2003photometric}. $\textbf{B}=(B_r, B_g, B_b)^T$ represents light brightness and $\mathcal{B}=B_r+B_g+B_b$. $\textbf{R}=(R_r,R_g,R_b)^T$ represents background reflection and $\mathcal{R}=R_r+R_g+R_b$. $\alpha=\textbf{B}/\mathcal{B}$ and $\pi=\textbf{R}/\mathcal{R}$ represents the chromaticities of $\textbf{B}$ and $\textbf{R}$, respectively. $T$ is the the exposure time and $t$ is the time for a raindrop to pass through pixel x. In the Eq.(\ref{eq:i}), the first term is the rain streak term and the second term is the background term. When employing any an existing color constancy algorithm to estimate $\alpha$, we can get the following normalization step:
\begin{equation}
\mathcal{O}({x})=\frac{\tilde{\mathcal{O}}({x})}{\alpha}={O}_{r s}({x}) \mathbf{i}+{O}_{b g}({x}),
\label{eq:rs}
\end{equation}
where $\mathbf{i}=(1,1,1)^T$, $O_{rs}=t \beta_{r s}\mathcal{B}$, and $O_{bg}=(T-t)\mathcal{R}/\alpha$. The vector division is  the element-wise division. When we normalize the image, the light chromaticity will be cancelled and the color effect of the spectral sensitivities will also be cancelled. Hence, based on Eq.(\ref{eq:rs}), given a rain image $\mathcal{O}$, the residue channel prior $\mathcal{P}$ of $\mathcal{O}$ can be defined as:
\begin{equation}
    \mathcal{P}(x)=\max _{c \in r, g, b}  \mathcal{O}^{c}(x)-\min _{d \in r, g, b}\mathcal{O}^{d}(x).
    \label{res_p}
\end{equation}
Due to the rain streak term in Eq.(\ref{eq:i}) is achromatic, whose values are canceled when employing color constancy, residue channel prior can be free from rain streaks, as shown in Fig.~\ref{fig:rcp}.  Moreover, according to Li~\etal~\cite{li2018robust}, without employing color constancy, the residue channel prior can still work. Since the dominant gray atmospheric light generated by a cloudy sky, the appearance of rain streaks is already achromatic in most cases. Based on this observation, RCP can extract a more complete and accurate object structure. Therefore, RCP has been introduced to the model for image deraining guidance.

\subsubsection{RCP Extraction Module} 
In order to extract high-dimensional features of the RCP, we propose an RCP extraction module. As shown in Fig.~\ref{fig:model}, a convolution is firstly used to obtain the initial feature maps $\mathcal{F}_{p}^{init}$ of $\mathcal{P}$. Furthermore, to reduce the noise in the initial features and enrich the semantic information of the feature, we use SE-ResBlock in Residual block (SRiR) to extract the deeper feature $\mathcal{F}_{p}$ of RCP.

\subsubsection{Interactive Fusion Module}

\begin{figure}
	\begin{center}
		\includegraphics[scale=0.48]{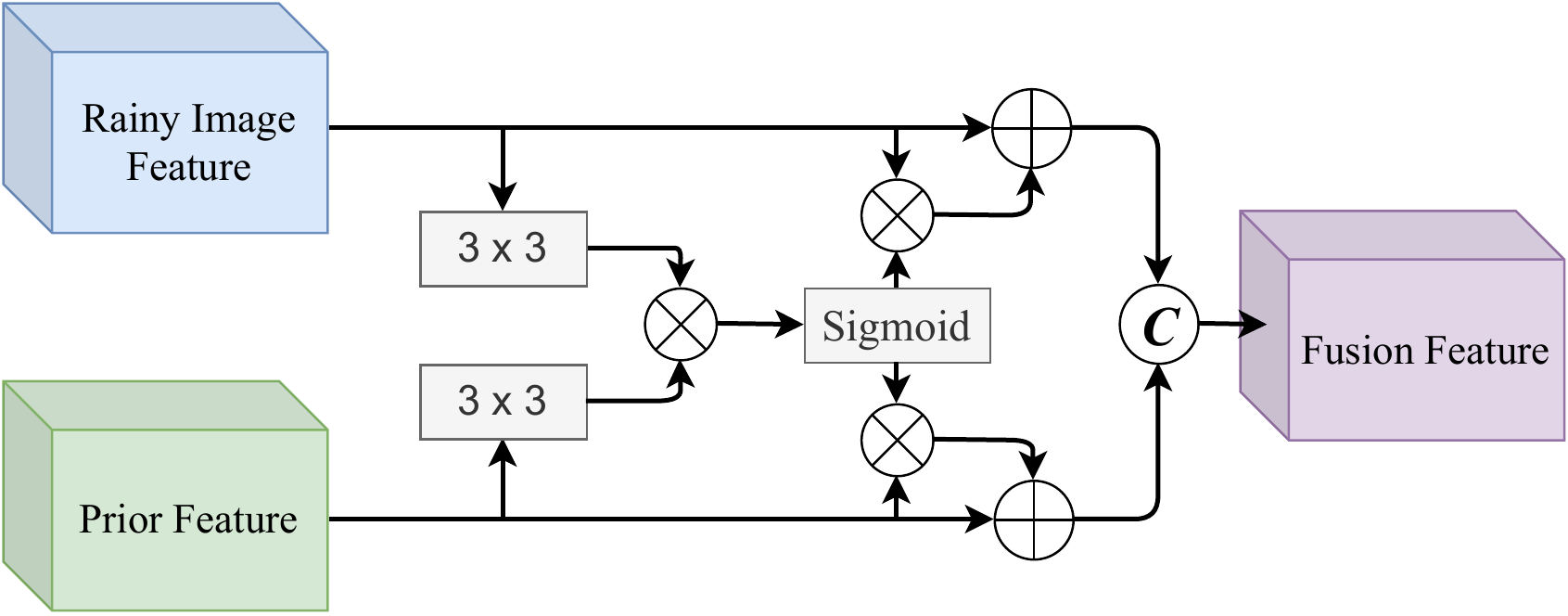}
	\end{center}
	\vspace{-10pt}
	\caption{The architecture of the propose Interactive Fusion Module (IFM).}
	\label{fig:ifm}
	\vspace{-10pt}
\end{figure}

Although RCP has clearer structural information than the rainy image, how to make full use of the RCP features to guide the model is still a challenging task. One simple solution is concatenating RCP features with image features together. However, this is not effective to guide the model deraining and may cause feature interference. To solve this problem, we propose an interactive fusion module (IFM) to progressively combine features together. As shown in Fig.~\ref{fig:ifm}, two convolution with $3 \times 3$ kernel size to map the rainy image feature $\mathcal{F}_o$ and the RCP feature $\mathcal{F}_{p}$ to $\widehat{\mathcal{F}}_o$ and $\widehat{\mathcal{F}}_p$. 
\begin{equation}
    \widehat{\mathcal{F}}_o = Conv(\mathcal{F}_o),
\end{equation}
\begin{equation}
    \widehat{\mathcal{F}}_p = Conv(\mathcal{F}_p),
\end{equation}

Next, we calculate the similarity map $\mathcal{S}$ between $\widehat{\mathcal{F}}_o$ and $\widehat{\mathcal{F}}_p$ by using element multiplication to enhance the background information of the rainy image destroyed by the rain streaks. Moreover, because the background of RCP is similar to the rainy image, the similarity map $\mathcal{S}$ can also highlight the feature information in the prior features, thereby further strengthening the  structure of the prior feature.
\vspace{-4pt}
\begin{equation}
    \mathcal{S} = Sigmoid(\widehat{\mathcal{F}}_o \otimes \widehat{\mathcal{F}}_p),
\end{equation}
\vspace{-12pt}
\begin{equation}
    \mathcal{F}_{o}^{s} = \mathcal{S} \otimes \mathcal{F}_o,
\end{equation}
\vspace{-12pt}
\begin{equation}
    \mathcal{F}_{p}^{s} = \mathcal{S} \otimes \mathcal{F}_p,
\end{equation}
where $\mathcal{F}_{o}^{s}$ denotes the activated feature of $\mathcal{F}_{o}$ and $\mathcal{F}_{p}^{s}$ denotes the activated feature of $\mathcal{F}_{p}$. Finally, the activated feature is add with the original feature and the output $\widetilde{\mathcal{F}}$ of IFM can be obtained by concatenating the features after addition:
\vspace{-10pt}
\begin{equation}
    \widetilde{\mathcal{F}}_{o} = \mathcal{F}_{o}^{s} + \mathcal{F}_o,
\end{equation}
\vspace{-12pt}
\begin{equation}
    \widetilde{\mathcal{F}}_{p} = \mathcal{F}_{p}^{s} + \mathcal{F}_p,
\end{equation}
\vspace{-12pt}
\begin{equation}
    \widetilde{\mathcal{F}} = Concat(\widetilde{\mathcal{F}}_{o},\widetilde{\mathcal{F}}_{p}),
\end{equation}

\subsubsection{Iterative Guidance Strategy}
\label{strategy}
\begin{figure}[t]
	\centering
	\begin{minipage}[c]{0.127\textwidth}
		\includegraphics[scale=0.166]{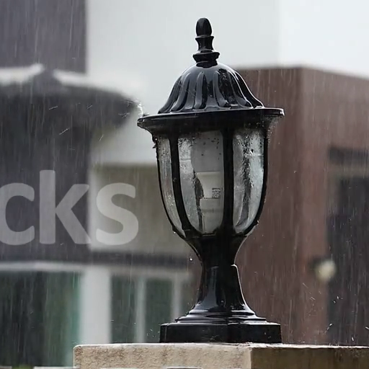}
		\includegraphics[scale=0.166]{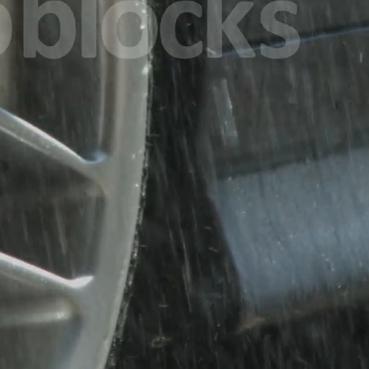}
		 \centerline{(a)}
	\end{minipage}
    \begin{minipage}[c]{0.127\textwidth}
		\includegraphics[scale=0.23]{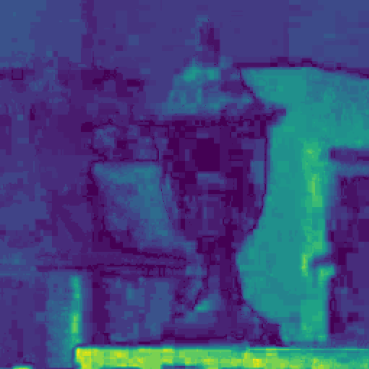}
		\includegraphics[scale=0.23]{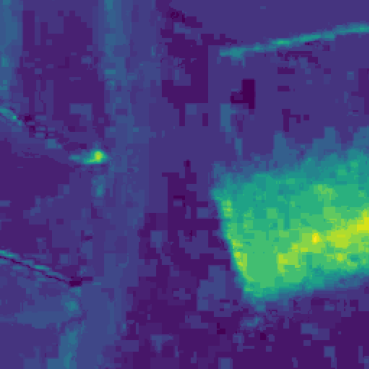}
		\centerline{(b)}
	\end{minipage}
	\begin{minipage}[c]{0.127\textwidth}
		\includegraphics[scale=0.23]{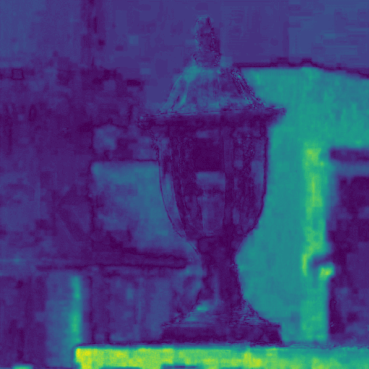}
		\includegraphics[scale=0.23]{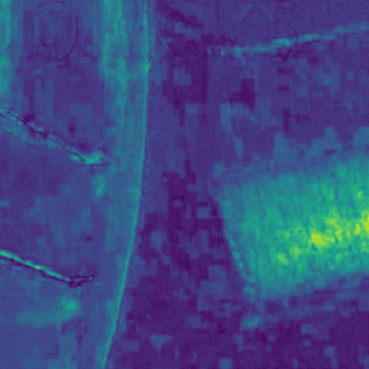}
		\centerline{(c)}
	\end{minipage}
	\vspace{-10pt}
	\caption{Comparison between the RCP of rainy images and the RCP of output results. (a) is rainy images, (b) is the RCP of rainy images, and (c) is the RCP of output results. It is  obvious observed that the structure of RCP of output results is more obvious than rainy images.}
	\label{fig:prior_compare}
	\vspace{-10pt}
\end{figure}

As shown in Fig.~\ref{fig:prior_compare}, with the improvement of image quality, the extracted RCP also shows better results.
Based on this observation, we propose an iterative guidance strategy to obtain a clearer RCP and replace the RCP of rainy images, that is, a rain removal result $\mathcal{B}_n$ can be obtained from the ${WMLM}_n$ output feature $\mathcal{F}_n$ and the clearer structure of prior $\mathcal{P}_{n+1}$ can be get from the $\mathcal{B}_n$.
\vspace{-2pt}
\begin{equation}
    \left\{\begin{array}{l}
\mathcal{F}_n = {WMLM}_n (\mathcal{\widehat{F}}_n, \mathcal{P}_{n}; \mathcal{\theta}_n),\\
\mathcal{B}_{n} = Conv(\mathcal{F}_n),\\
\mathcal{P}_{n+1} = REM(\mathcal{B}_n), \quad if\,\, n=1,2,\\
\end{array}\right.
\vspace{-2pt}
\end{equation}
where the $\mathcal{\theta}_n$ represents the weights of ${WMLM}_n$, $Conv$ stands of the output convolution, $REM$ means RCP extraction module, and $n$ is the number of ${WMLM}$. In our proposed model, $n$ is set to 3. The $\mathcal{\widehat{F}}_n$ is input feature map of the $WMLM_n$. 
Moverover, to enable the model to learn richer features and generate better results, we leverage the idea of \textit{Ensemble Learning} to adopt the concat operation to fuse the output features of the previous WMLM with the current WMLM output feature. \label{ensemble}

\subsection{Loss Function}

We employ $\mathcal{L}_2$ loss as our objective function. As mentioned above, under the iterative update strategy, the model will output three results. Thus, the comprehensive loss function of our proposed model can be formulated as:
\begin{equation}
    \mathcal{L} = \sum_{i}\left\|\mathcal{B}_{i}-\hat{\mathcal{B}}\right\|^{2}, \quad i=1,2,3.
\end{equation}
Where, $\hat{\mathcal{B}}$ denotes the rain-free image(GT) and $\mathcal{B}_{i}$ denotes output results in different stage.

\begin{table*}[htbp]
	\centering
	\setlength{\tabcolsep}{1.5mm}
	\renewcommand{\arraystretch}{0.9}
		\centering
			\begin{tabular}{ccccccccccccc}
				\toprule
				\multirow{2}{*}{Methods}& \multirow{2}{*}{Param} & \multirow{2}{*}{Time} &
				\multicolumn{2}{c}{Rain200L}&\multicolumn{2}{c}{Rain200H}&\multicolumn{2}{c}{Rain800}&\multicolumn{2}{c}{Rain1200}&\multicolumn{2}{c}{SPA-Data}\cr
				\cmidrule(lr){4-5} \cmidrule(lr){6-7} \cmidrule(lr){8-9} \cmidrule(lr){10-11} \cmidrule(lr){12-13}
				       &  & $128\times128$ & PSNR & SSIM & PSNR & SSIM & PSNR & SSIM & PSNR & SSIM & PSNR & SSIM\cr
				\midrule
				GMM\cite{li2016rain}    & ----- & 27.961s & 28.66 & 0.8652 & 14.50 & 0.4164 & 25.71 & 0.8020 & 25.81 & 0.8344 & 34.30 & 0.9428\cr
				DSC\cite{luo2015removing}    & ----- & 7.947s & 27.16 & 0.8663 & 14.73 & 0.3815 & 22.61 & 0.7530 & 24.24 & 0.8279 & 34.95 & 0.9416\cr
				DDN\cite{fu2017removing}    & 0.06M & 0.278s & 34.68 & 0.9671 & 26.05 & 0.8056 & 25.87 & 0.8018 & 30.97 & 0.9116 & 36.16 & 0.9463\cr
				RESCAN\cite{li2018recurrent} & 0.15M & 0.016s & 36.09 & 0.9697 & 26.75 & 0.8353 & 26.58 & 0.8726 & 33.38 & 0.9417 & 38.11 & 0.9707\cr
				PReNet\cite{ren2019progressive} & 0.17M & 0.012s & 37.70 & 0.9842 & 29.04 & 0.8991 & 27.06 & 0.9026 & 33.17 & 0.9481 & 40.16 & 0.9816\cr
				DCSFN\cite{wang2020dcsfn}  & 6.45M & 0.253s & 39.37 & 0.9854 & 29.25 & 0.9075 & 28.38 & 0.9072 & \underline{34.31} & \underline{0.9545} & ----- & -----\cr
				DRDNet\cite{deng2020detail} & 2.72M & 0.069s & 39.05 & 0.9862 & 29.15 & 0.8921 & 28.21 & 0.9012 & 34.02 & 0.9515 & 40.89 & 0.9784\cr
				RCDNet\cite{wang2020model} & 3.17M & 0.068s & \underline{39.87} & \underline{0.9875} & \underline{30.24} & \underline{0.9098}  & \underline{28.59} & \underline{0.9137} & 34.08 & 0.9532 & \underline{41.47} & \underline{0.9834}\cr
				SPDNet(Ours)                       & 3.04M & 0.055s & \textbf{40.59} & \textbf{0.9880} & \textbf{31.30} & \textbf{0.9217} & \textbf{30.21} & \textbf{0.9152} & \textbf{34.57} & \textbf{0.9561} & \textbf{43.55} & \textbf{0.9875}\cr
				\bottomrule
			\end{tabular}
			\caption{Quantitative experiments evaluated on four recognized synthetic datasets. The best and the second best results have been boldfaced and underlined.}
			\label{Tab:compare}
			\vspace{-10pt}
\end{table*}
\begin{figure*}[t]
	\centering
	\begin{minipage}[c]{0.1\textwidth}
		\includegraphics[scale=0.105]{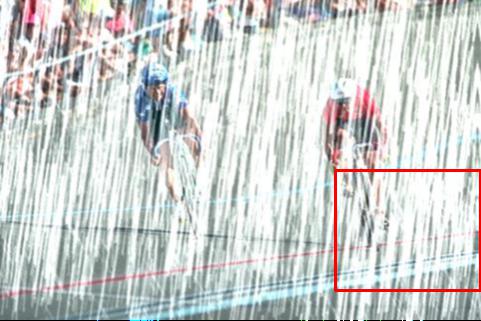}
		\includegraphics[scale=0.105]{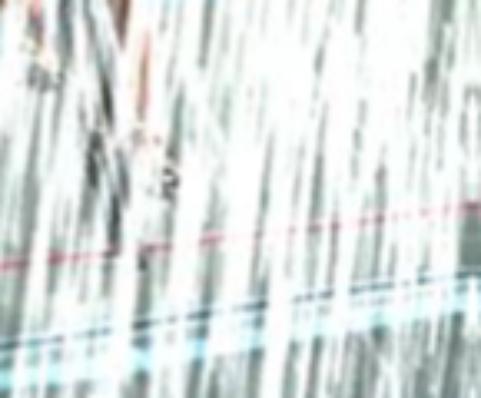}
		\small \centerline{Rainy images}
	\end{minipage}
	\begin{minipage}[c]{0.1\textwidth}
		\includegraphics[scale=0.105]{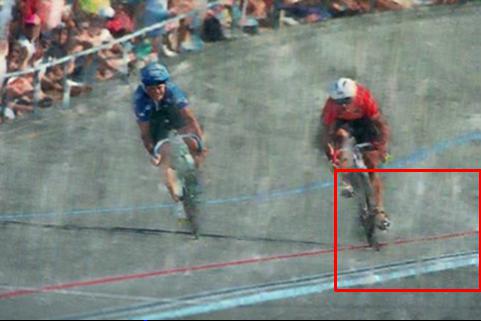}
		\includegraphics[scale=0.105]{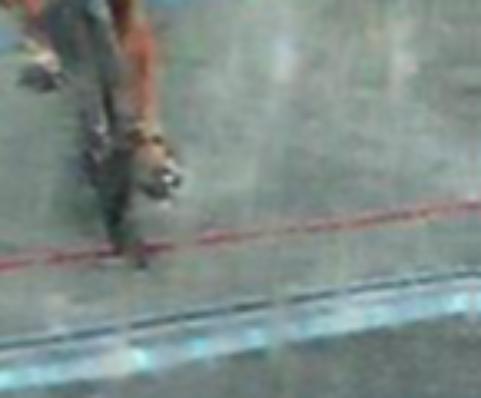}
		\centerline{DDN}
	\end{minipage}
	\begin{minipage}[c]{0.1\textwidth}
		\includegraphics[scale=0.105]{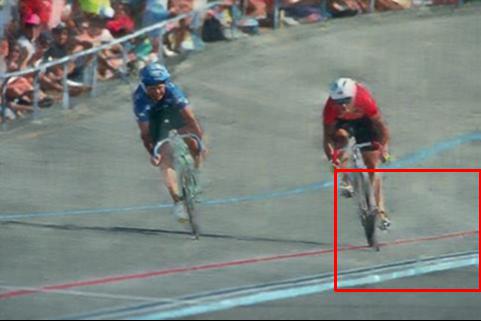}
		\includegraphics[scale=0.105]{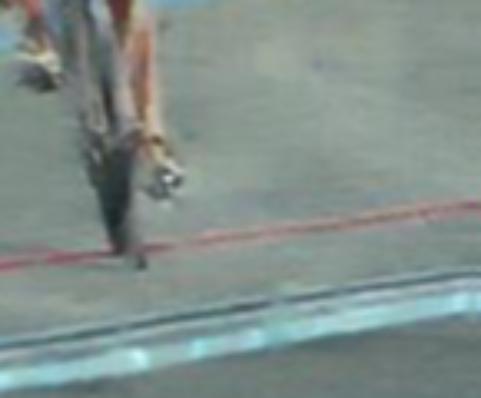}
		\centerline{RESCAN}
	\end{minipage}
	\begin{minipage}[c]{0.1\textwidth}
		\includegraphics[scale=0.105]{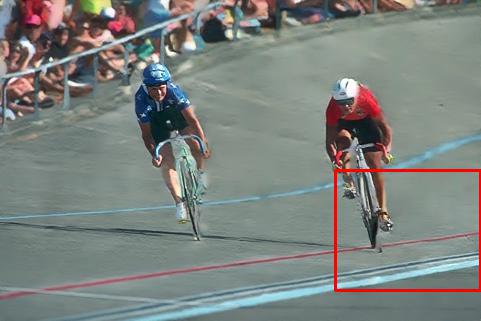}
		\includegraphics[scale=0.105]{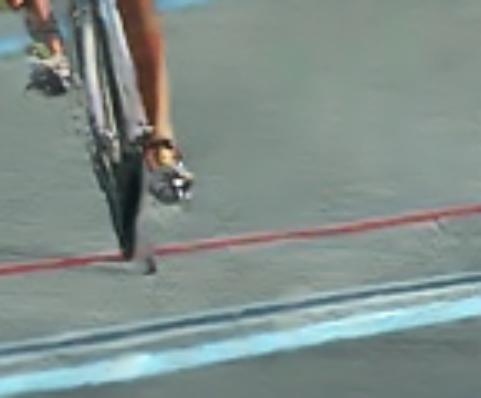}
		\centerline{PReNet}
	\end{minipage}
	\begin{minipage}[c]{0.1\textwidth}
		\includegraphics[scale=0.105]{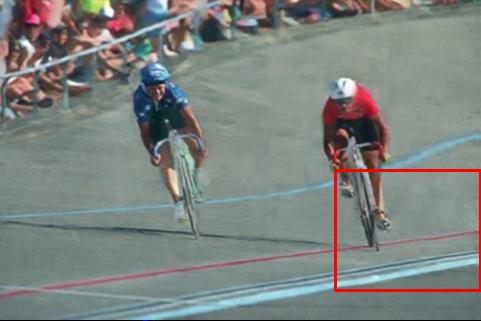}
		\includegraphics[scale=0.105]{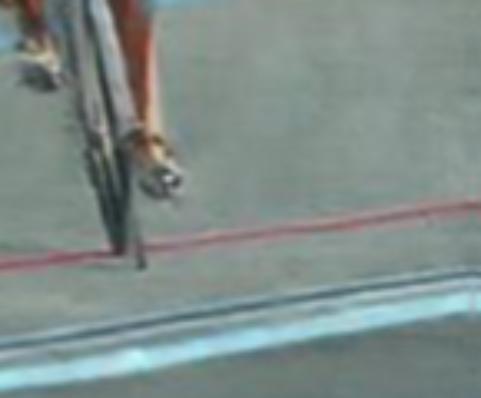}
		\centerline{DRDNet}
	\end{minipage}
	\begin{minipage}[c]{0.1\textwidth}
		\includegraphics[scale=0.105]{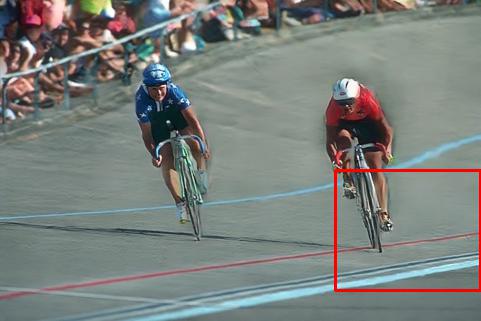}
		\includegraphics[scale=0.105]{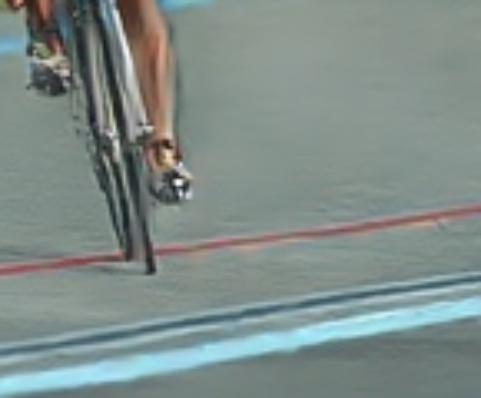}
		\centerline{DCSFN}
	\end{minipage}
	\begin{minipage}[c]{0.1\textwidth}
		\includegraphics[scale=0.105]{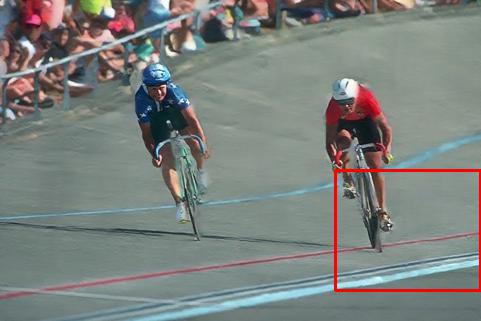}
		\includegraphics[scale=0.105]{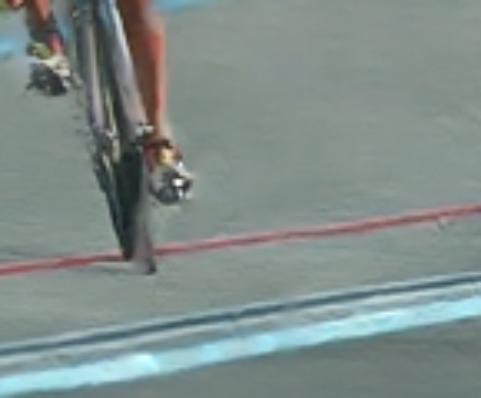}
		\centerline{RCDNet}
	\end{minipage}
	\begin{minipage}[c]{0.1\textwidth}
		\includegraphics[scale=0.105]{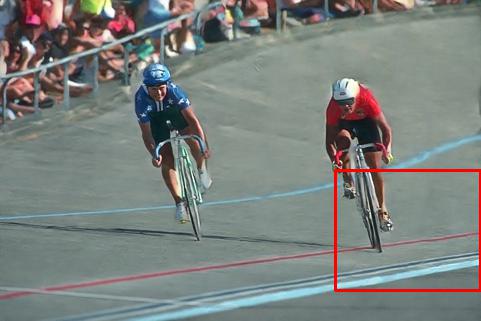}
		\includegraphics[scale=0.105]{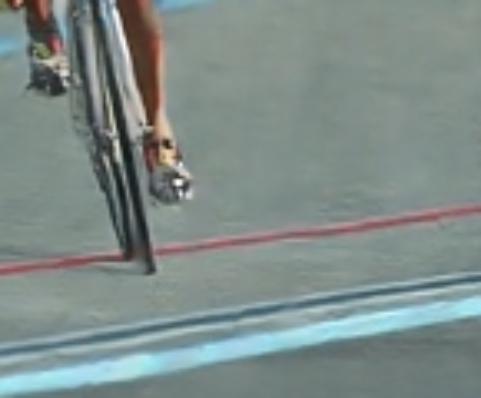}
		\centerline{SPDNet}
	\end{minipage}
	\begin{minipage}[c]{0.1\textwidth}
		\includegraphics[scale=0.105]{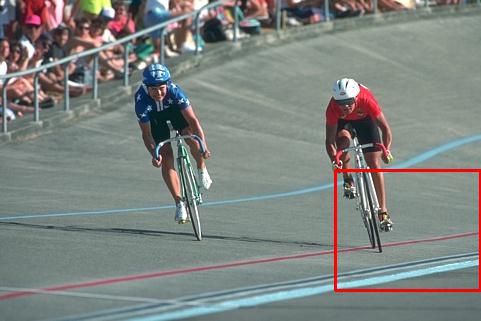}
		\includegraphics[scale=0.105]{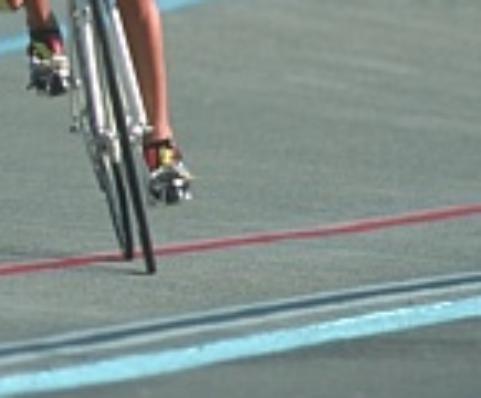}
		\centerline{GT}
	\end{minipage}
	\caption{Image deraining results tested in the synthetic datasets. The first row is rainy image, the output of different methods, and GT. The second row is the zoom results of the red window. It is obvious that SPDNet can reconstruct rain-free image with clearer structure.}
	\label{fig:34}
	\vspace{-12pt}
\end{figure*}
\begin{figure*}
	\centering
	\begin{minipage}[c]{0.11\textwidth}
		\includegraphics[scale=0.043]{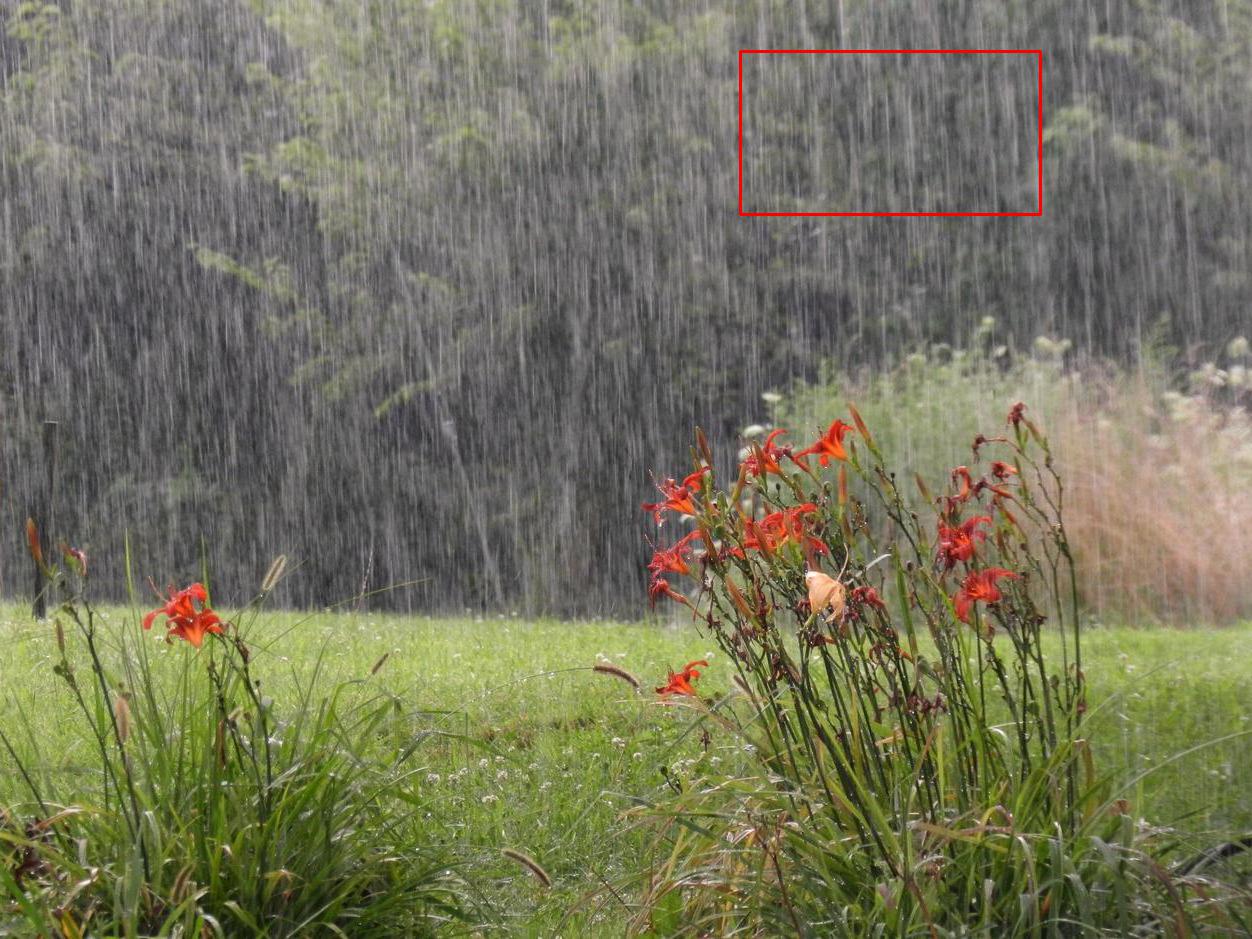}
		\includegraphics[scale=0.043]{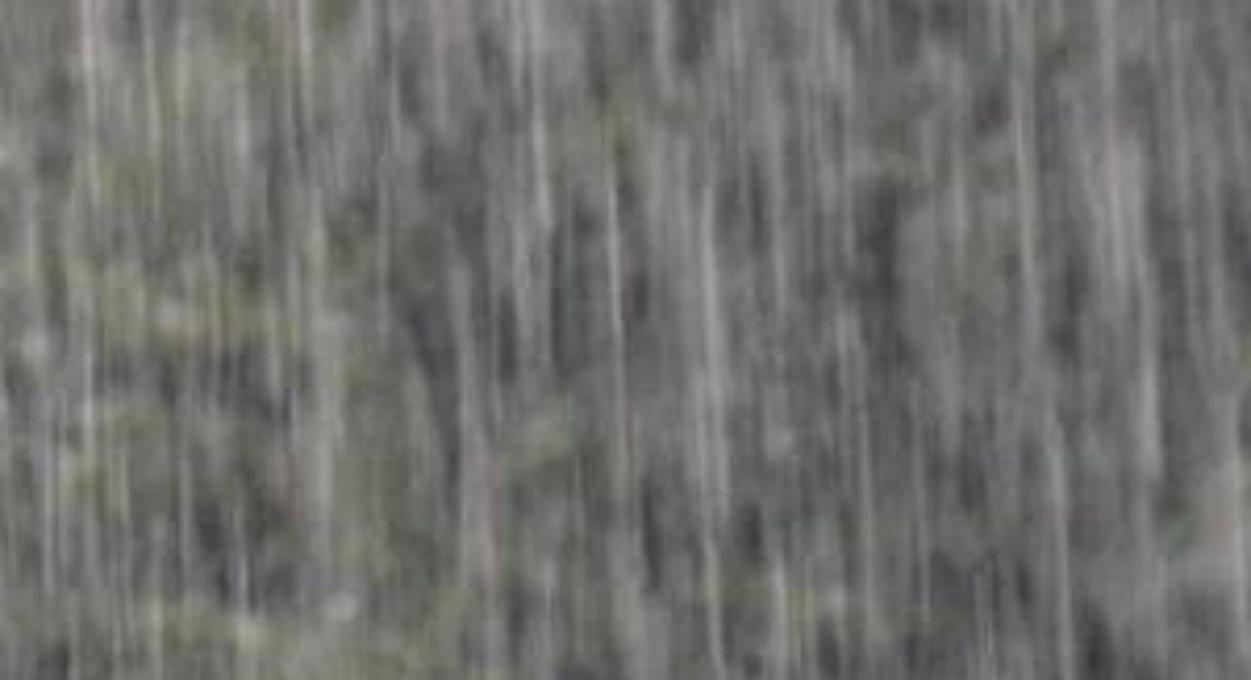}
		\includegraphics[scale=0.112]{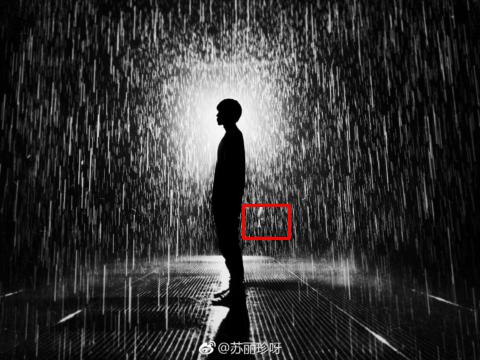}
		\includegraphics[scale=0.112]{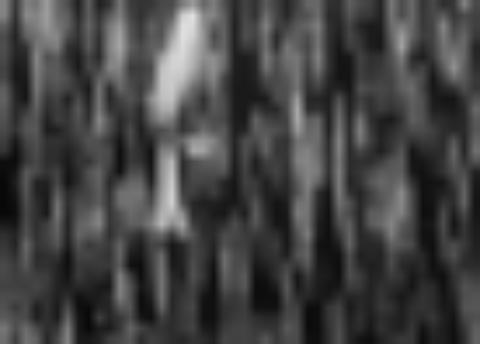}
		\small \centerline{Rainy image}
	\end{minipage}
	\begin{minipage}[c]{0.11\textwidth}
		\includegraphics[scale=0.043]{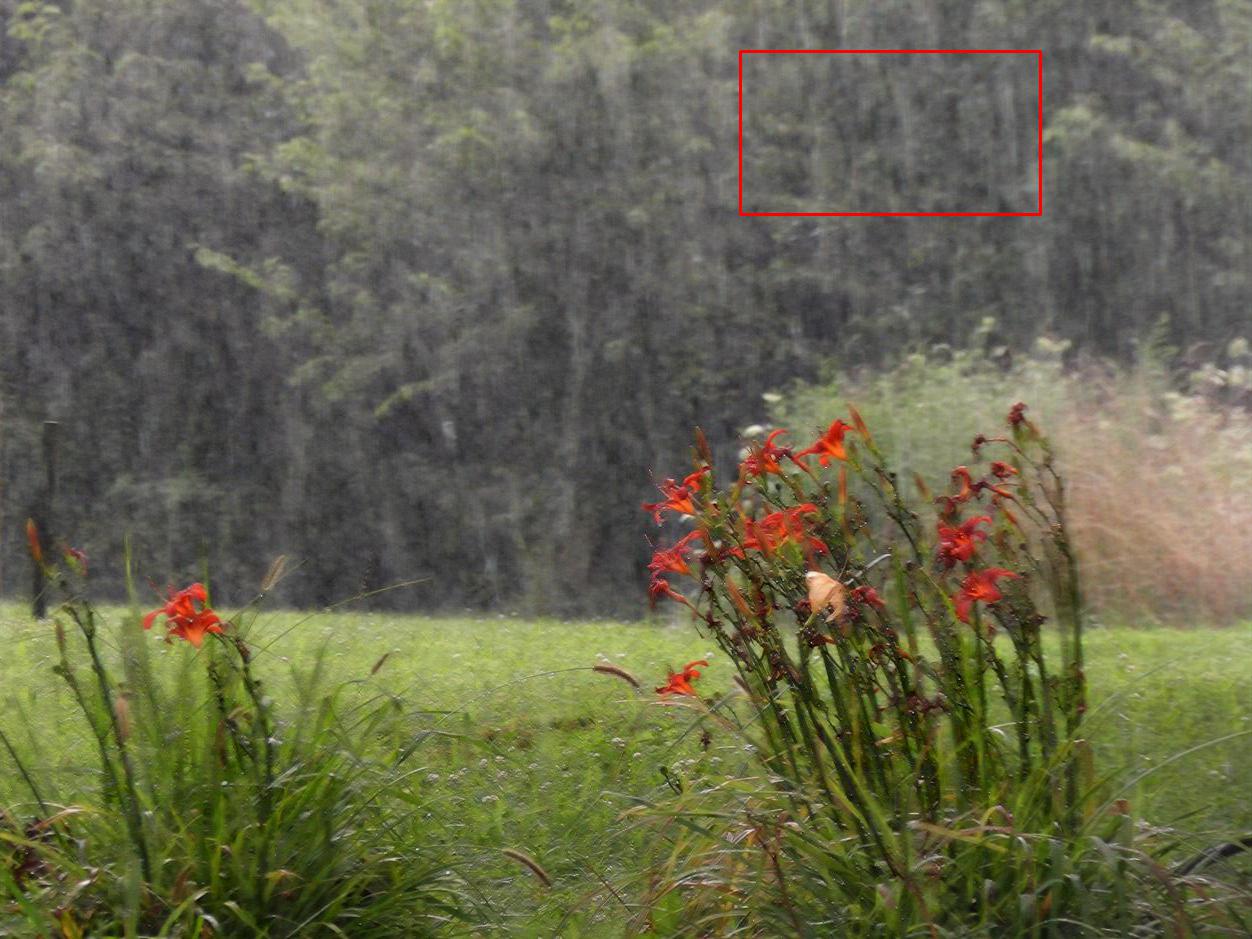}
		\includegraphics[scale=0.043]{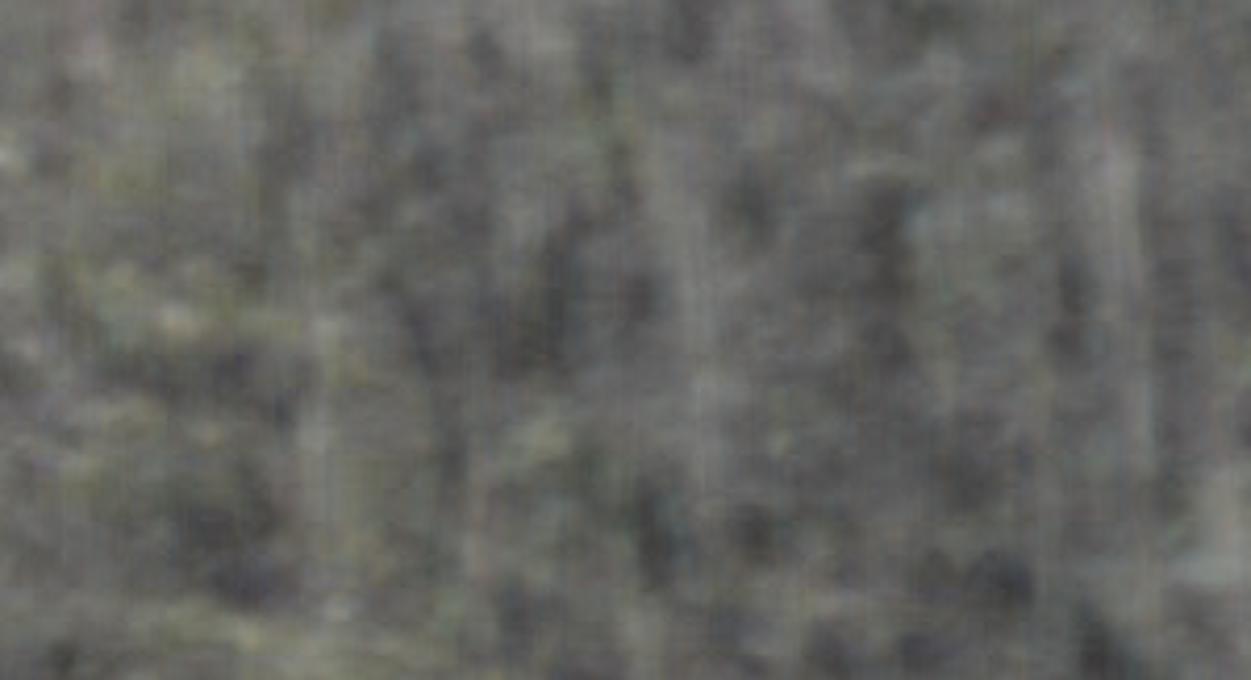}
		\includegraphics[scale=0.112]{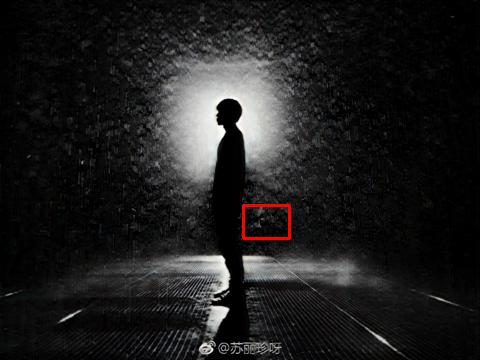}
		\includegraphics[scale=0.112]{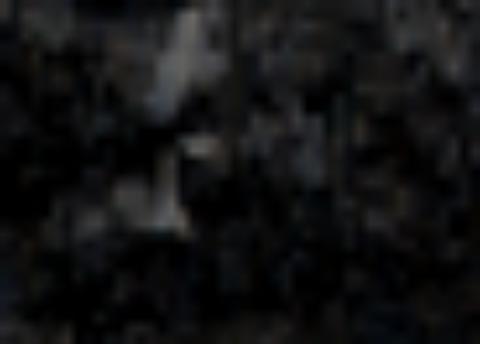}
		\centerline{DDN}
	\end{minipage}
	\begin{minipage}[c]{0.11\textwidth}
		\includegraphics[scale=0.043]{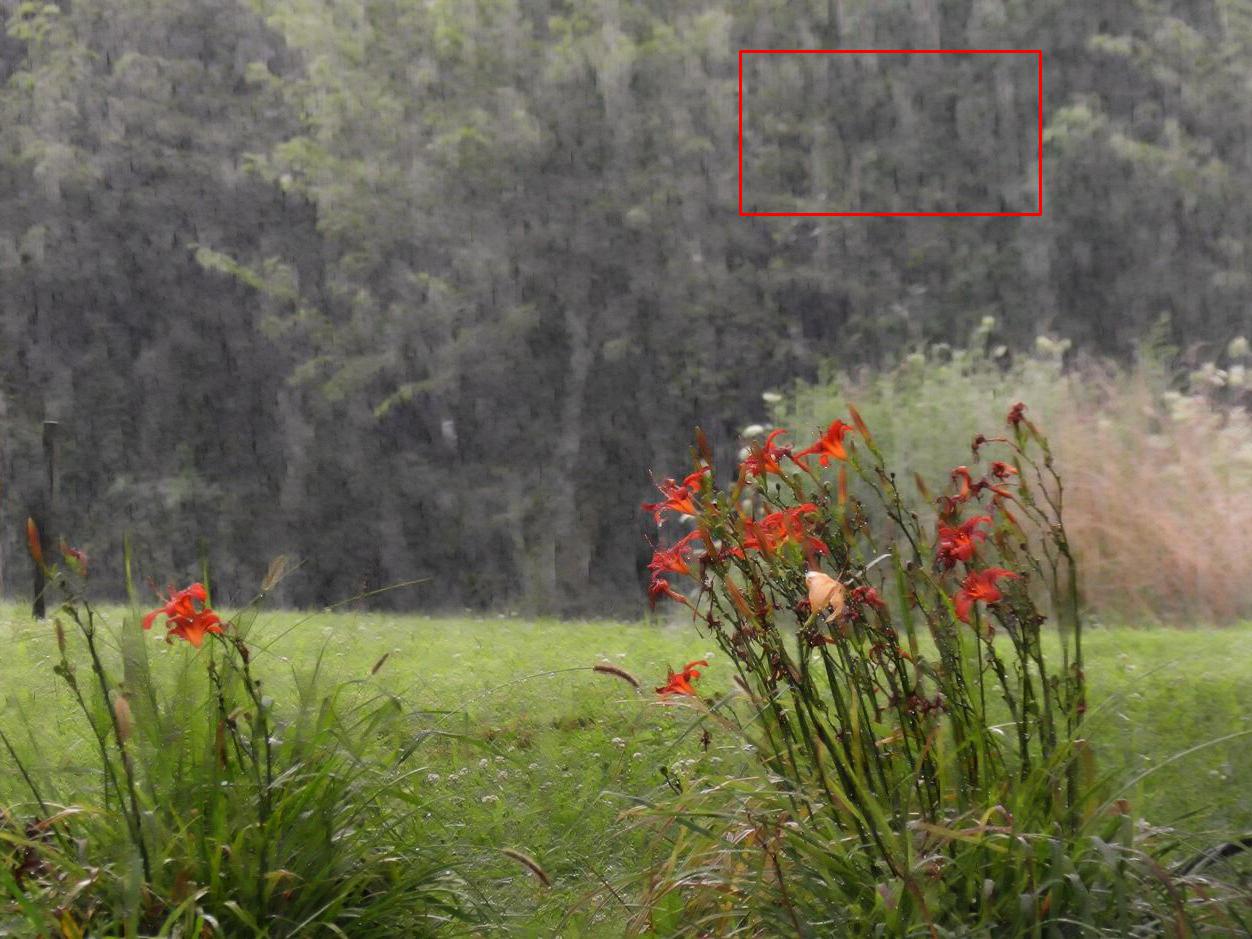}
		\includegraphics[scale=0.043]{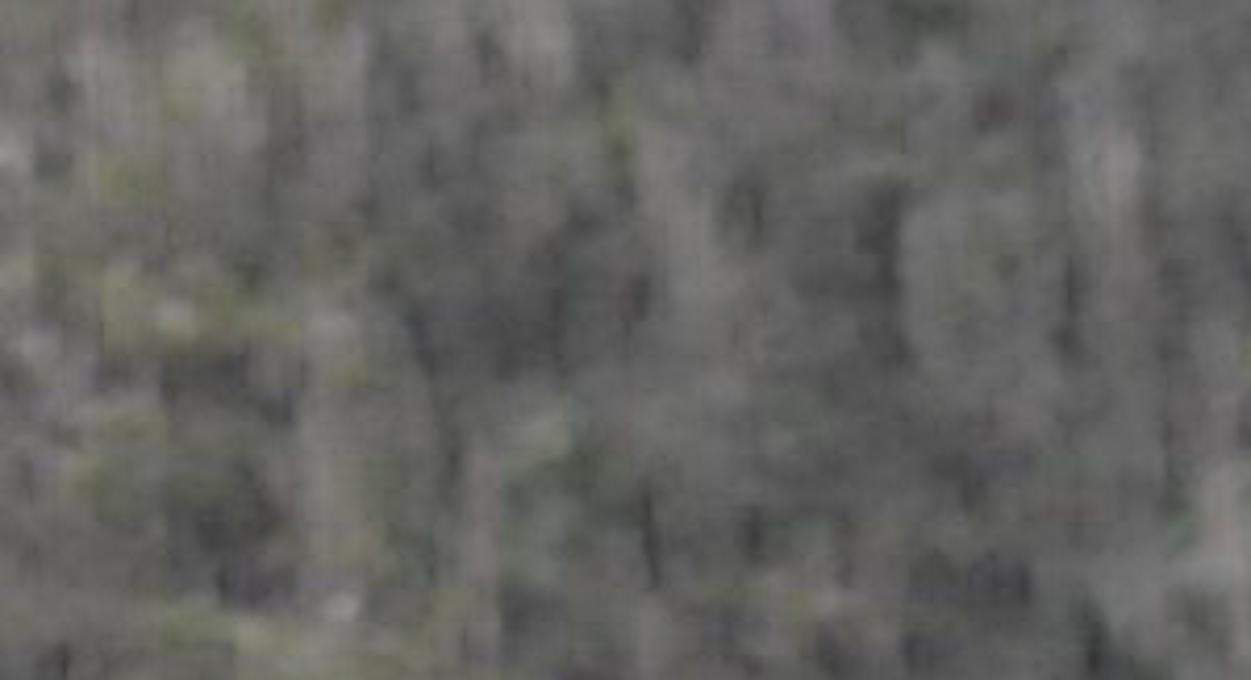}
		\includegraphics[scale=0.112]{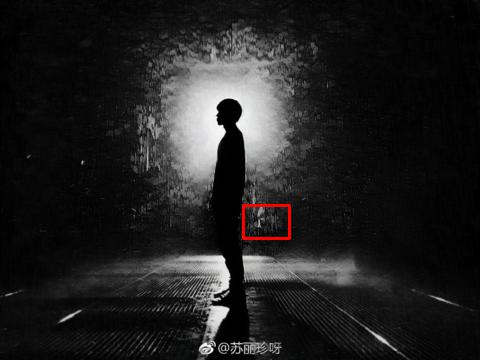}
		\includegraphics[scale=0.112]{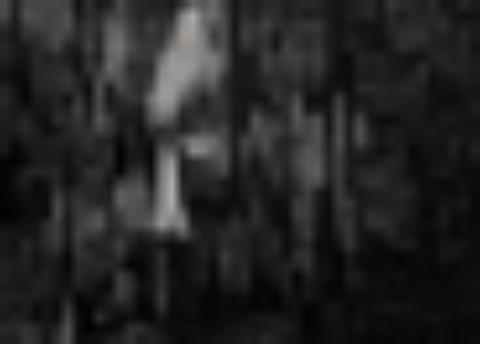}
		\centerline{RESCAN}
	\end{minipage}
	\begin{minipage}[c]{0.11\textwidth}
		\includegraphics[scale=0.043]{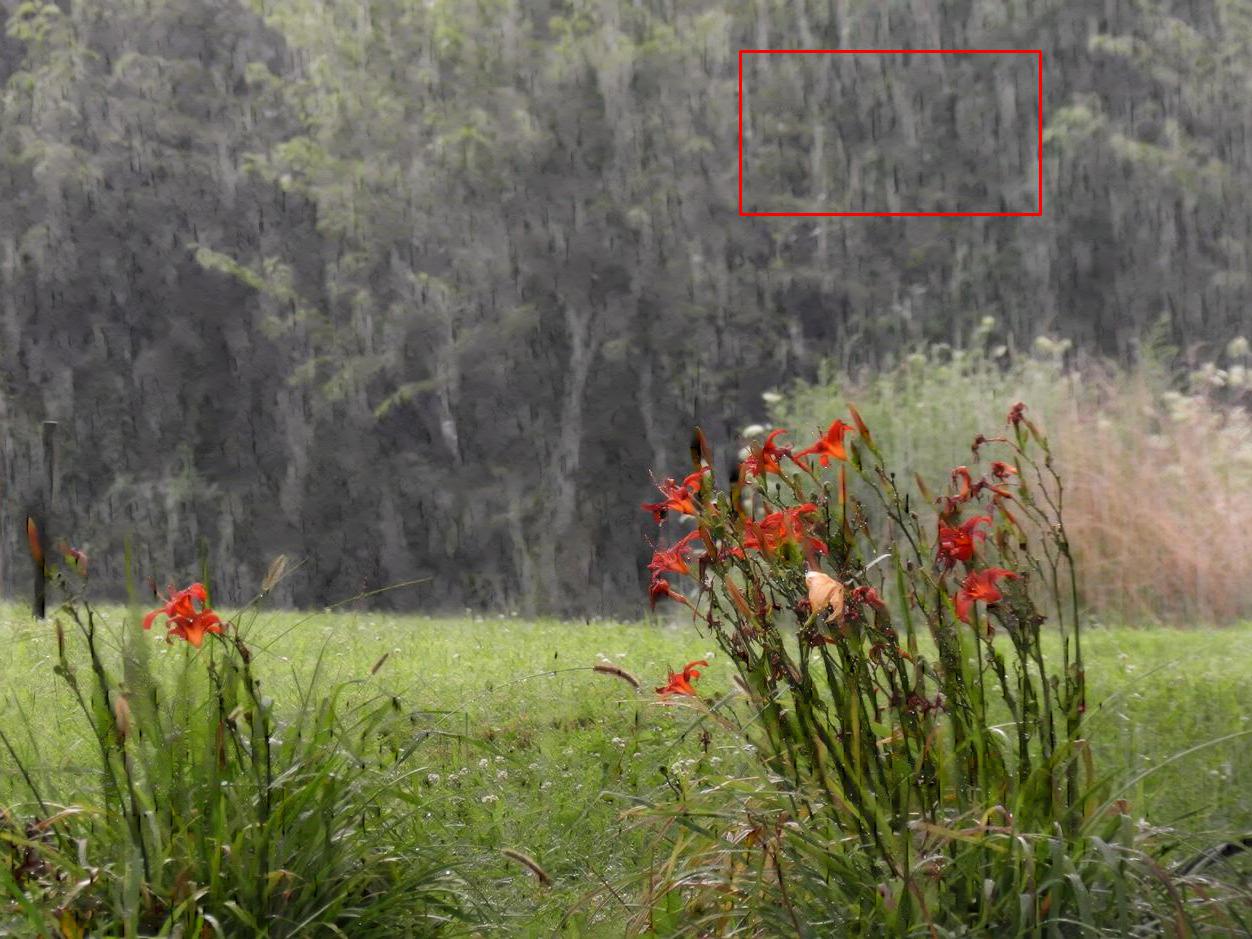}
		\includegraphics[scale=0.043]{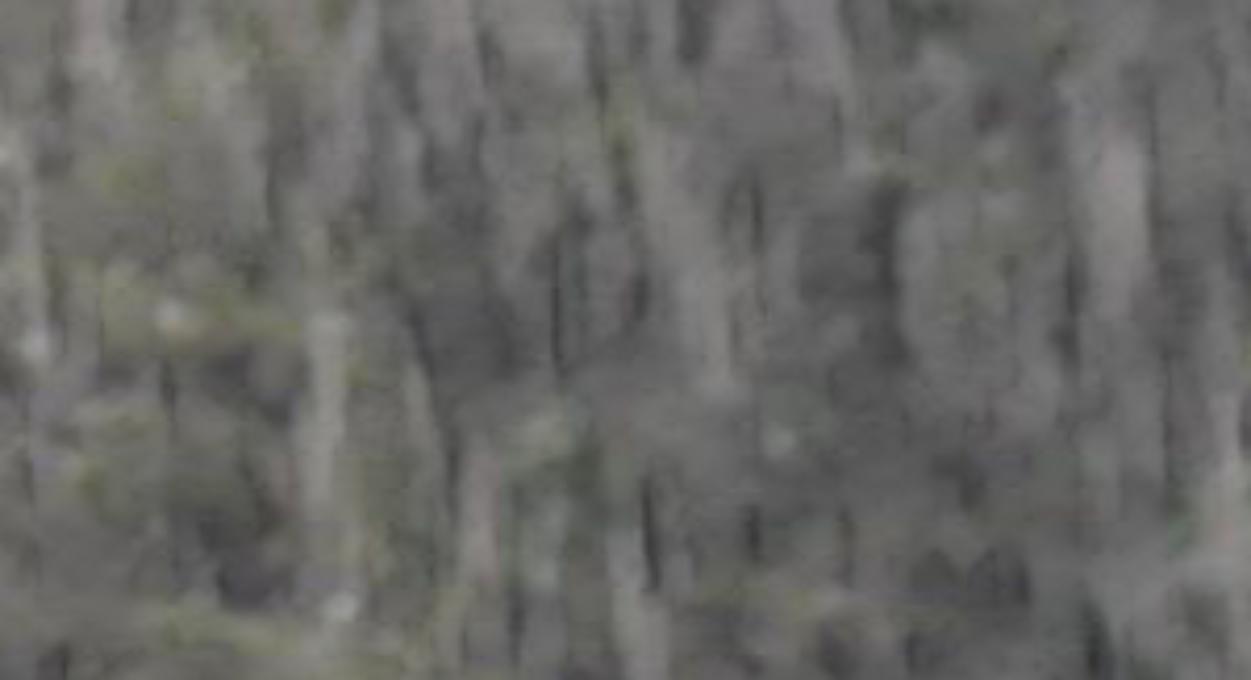}
		\includegraphics[scale=0.112]{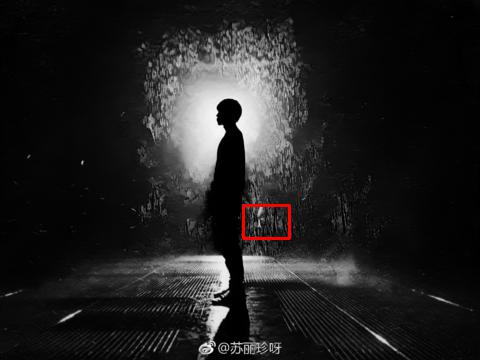}
		\includegraphics[scale=0.112]{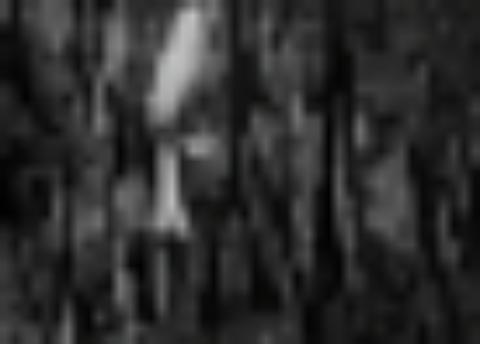}
		\centerline{PReNet}
	\end{minipage}
	\begin{minipage}[c]{0.11\textwidth}
		\includegraphics[scale=0.043]{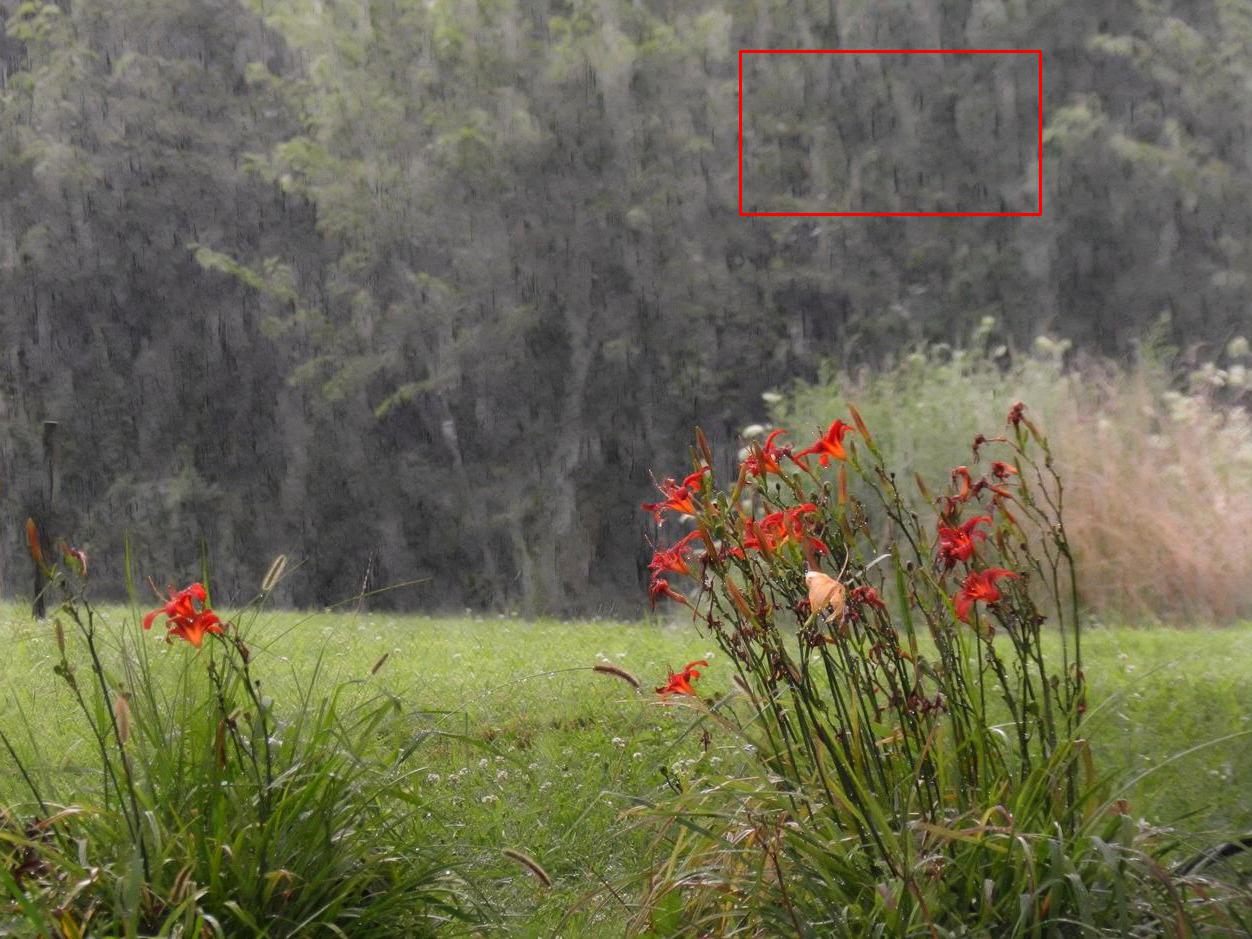}
		\includegraphics[scale=0.043]{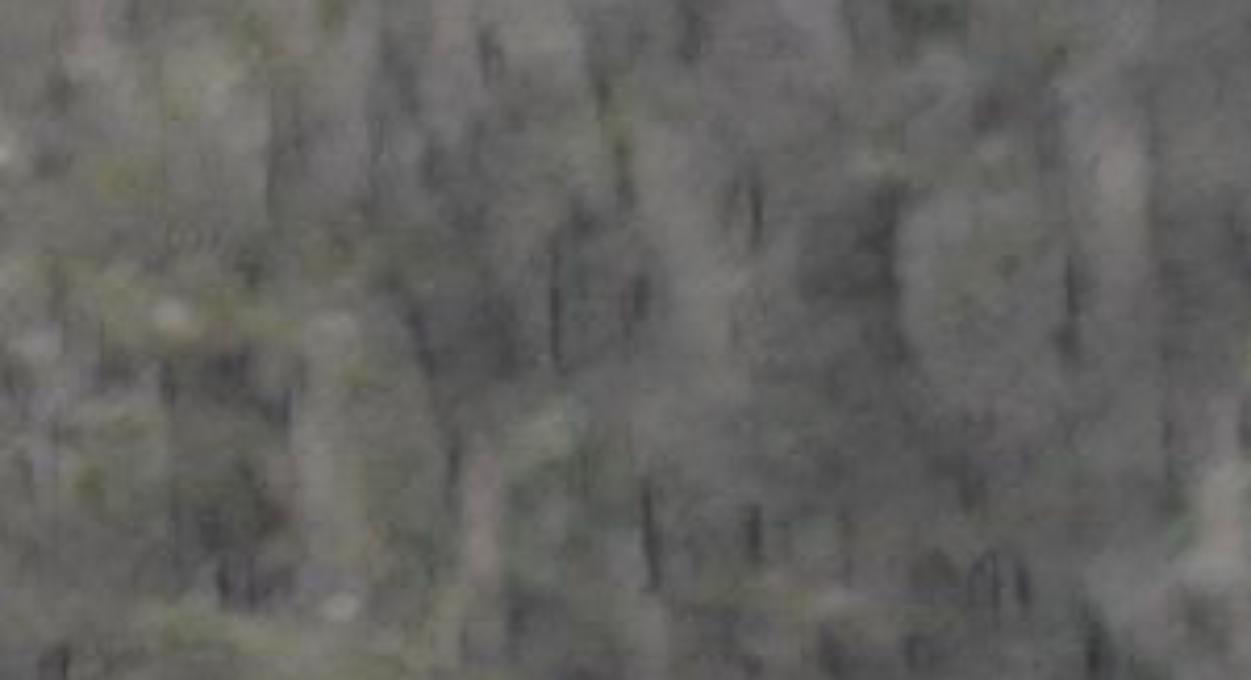}
		\includegraphics[scale=0.112]{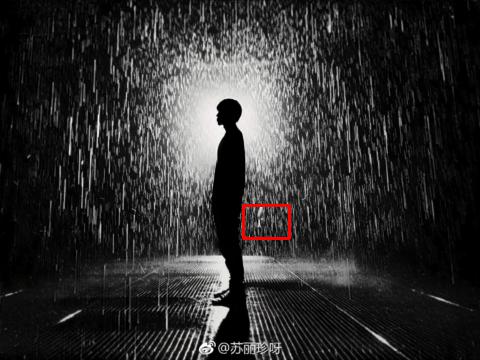}
		\includegraphics[scale=0.112]{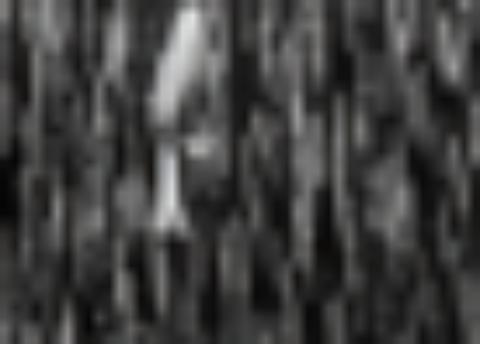}
		\centerline{DRDNet}
	\end{minipage}
	\begin{minipage}[c]{0.11\textwidth}
		\includegraphics[scale=0.043]{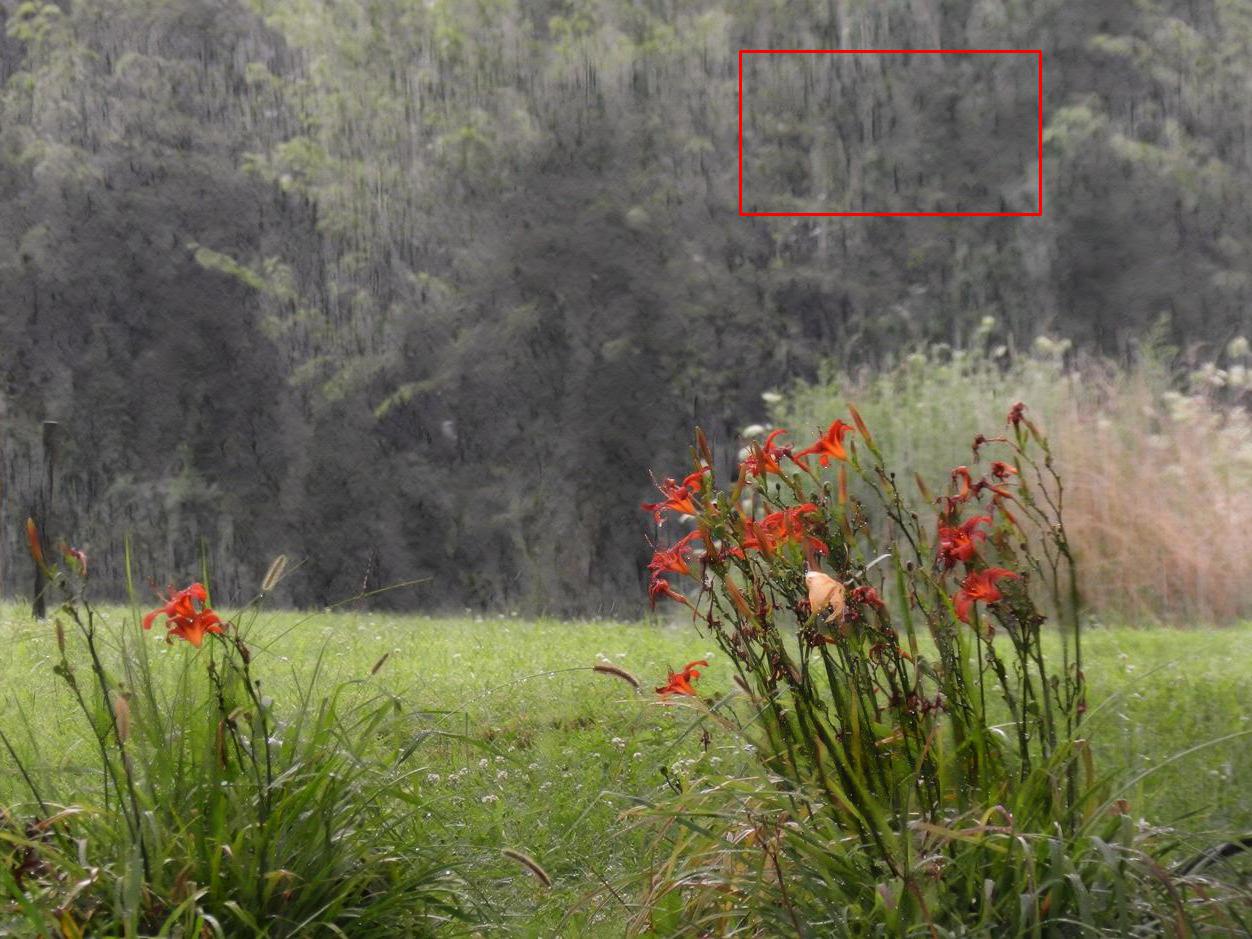}
		\includegraphics[scale=0.043]{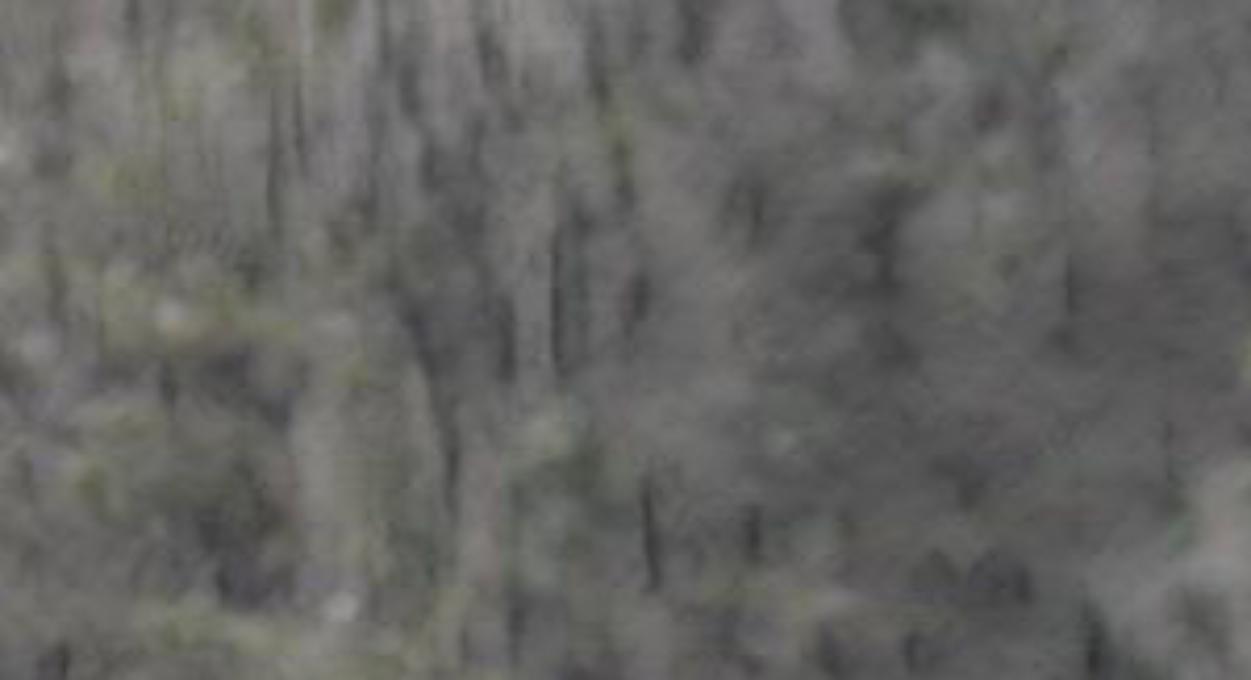}
		\includegraphics[scale=0.112]{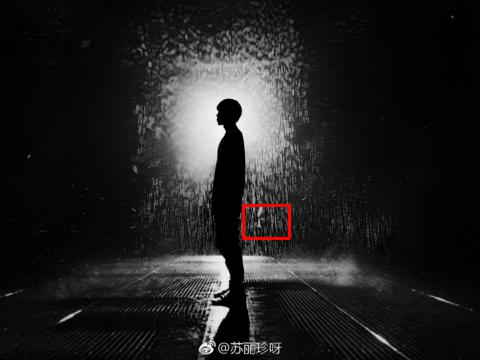}
		\includegraphics[scale=0.112]{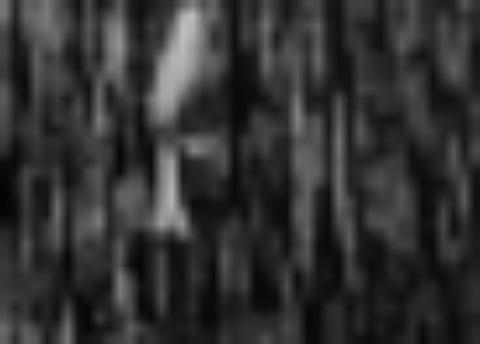}
		\centerline{DCSFN}
	\end{minipage}
	\begin{minipage}[c]{0.11\textwidth}
		\includegraphics[scale=0.043]{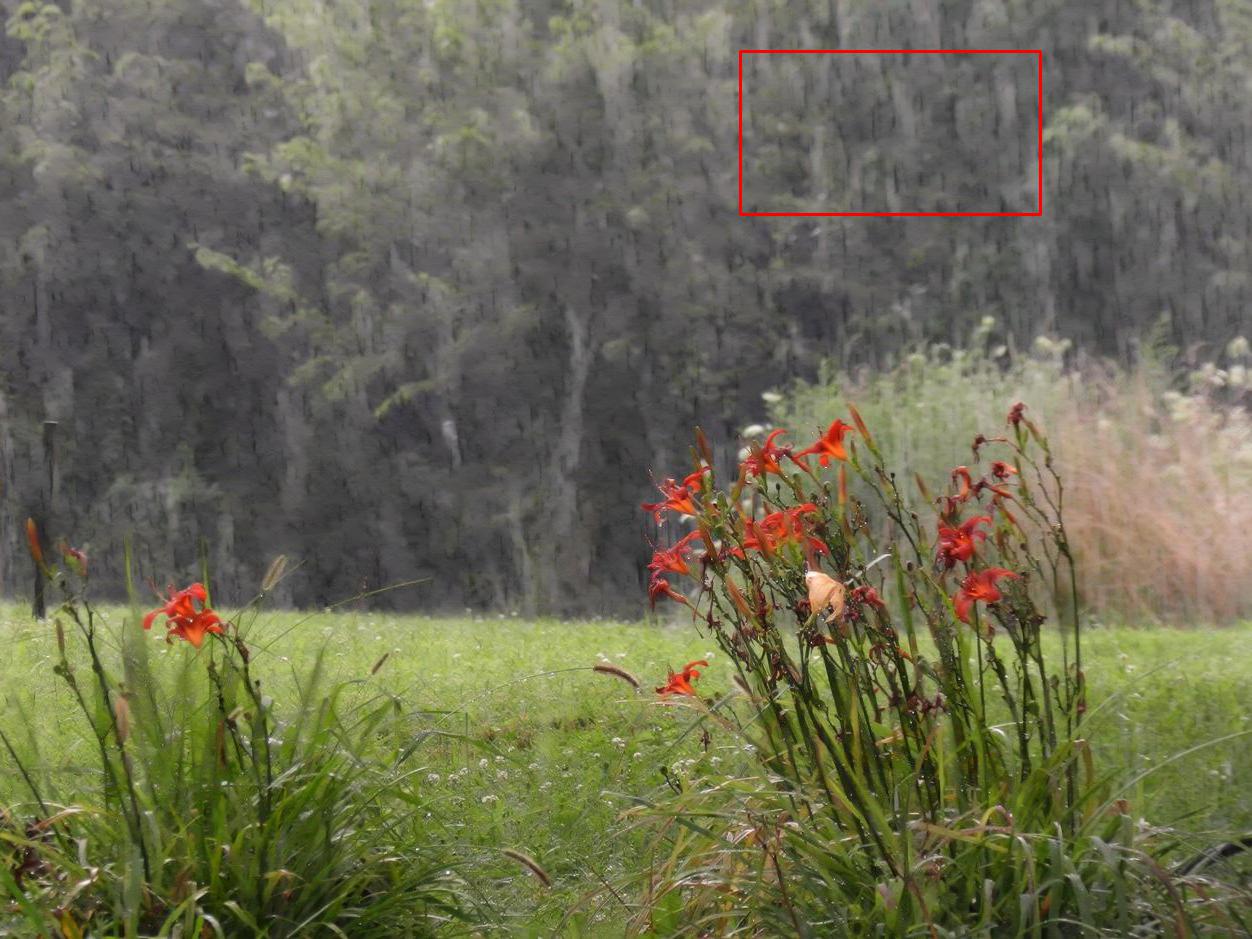}
		\includegraphics[scale=0.043]{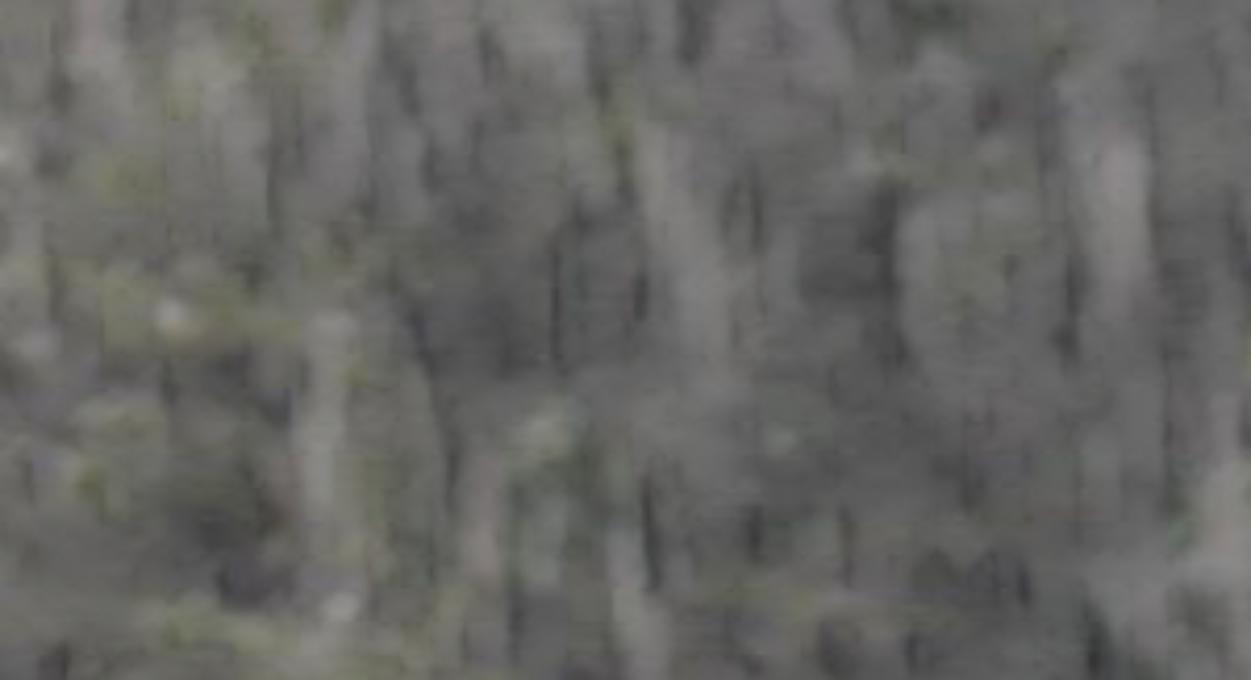}
		\includegraphics[scale=0.112]{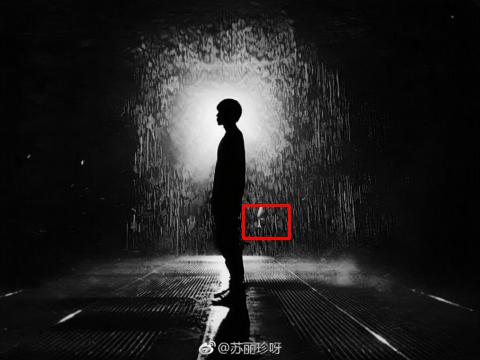}
		\includegraphics[scale=0.112]{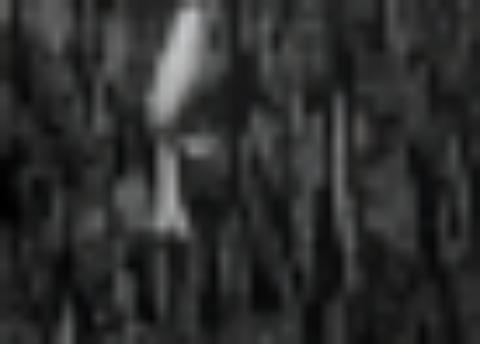}
		\centerline{RCDNet}
	\end{minipage}
	\begin{minipage}[c]{0.11\textwidth}
		\includegraphics[scale=0.043]{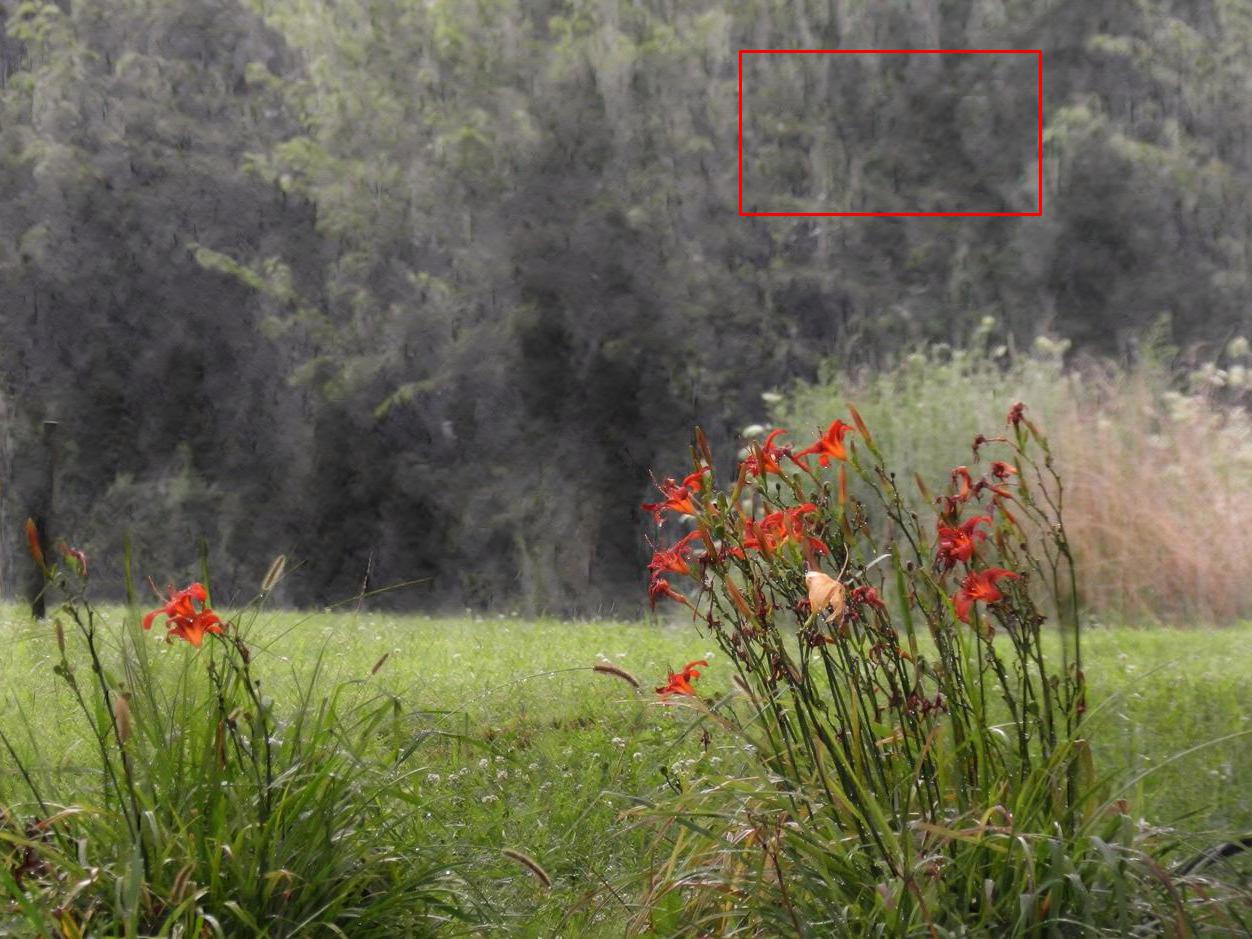}
		\includegraphics[scale=0.043]{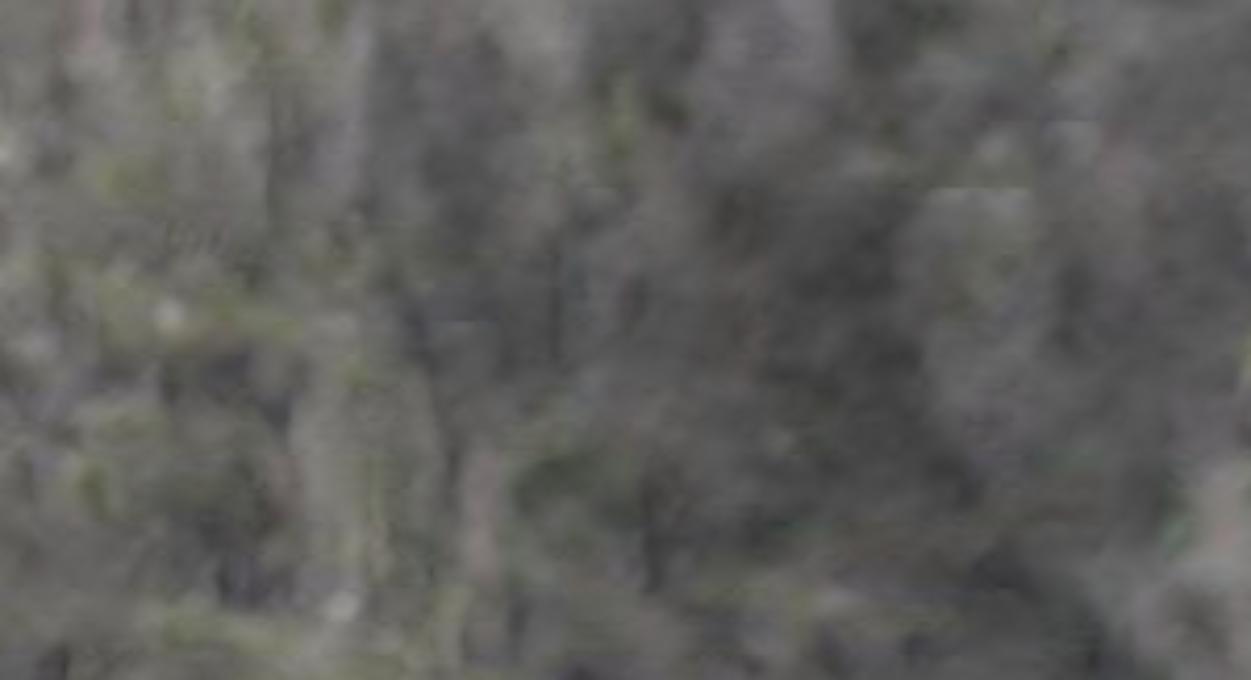}
		\includegraphics[scale=0.112]{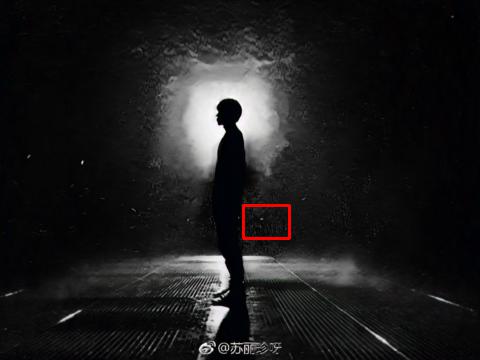}
		\includegraphics[scale=0.112]{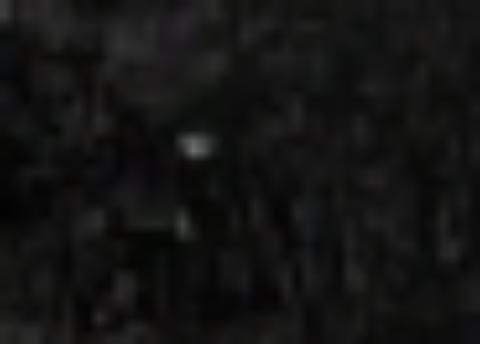}
		\centerline{SPDNet}
	\end{minipage}
	\caption{Image deraining results tested in the real-world dataset. The first row is rainy image and the output of different methods. The second row is the zoom results of the red window.}
	\label{fig:37}
	\vspace{-12pt}
\end{figure*}

\section{Experiment and Discussions}

\subsection{Datasets}

\textbf{Synthetic Datasets:} In our experiment, the \textit{Rain200L/H}, \textit{Rain800}, and \textit{Rain1200} are used to validate our proposed method. The \textit{Rain200L/H} are proposed by Yang~\etal~\cite{yang2017deep}, which consists of 1800 training images and 200 test images. Zhang~\etal~\cite{zhang2019image} collect a synthetic dataset named \textit{Rain800}, which contains 700 training images and 100 testing images. The \textit{Rain1200}~\cite{zhang2018density} consists of rainy images of different densities, including 12000 training sets and 1200 test sets.

\textbf{Real-world Datasets:} We use two real-word datasets to analyze the performance of all methods. The first one is the \textit{SPA-Data}~\cite{wang2019spatial},  which contains 638492 rainy/clear image pairs for training and the second one is supplied by Zhang~\etal~\cite{zhang2019image}, which does not contain clear images.  

\textbf{Implementation Details:} In our baseline, the number of WMLM is set to 3 and the number of RCP extraction module is set to 3. We use Adam optimizer with batch size of 16 and patch size of $128 \times 128$ for training on one NVIDIA Titan Xp GPU. The learning rate is set to $5 \times {10}^{-4}$ and we train the model for 300 epochs for \textit{Rain200L/H}, \textit{Rain800}, and \textit{Rain1200} datasets. For \textit{SPA-Data}, we train the model for 6 eopchs and the learning rate is set to $5 \times {10}^{-4}$.

\subsection{Comparison with the State-of-the-Arts}

\subsubsection{Synthesized Images}

\begin{table}
    \setlength{\tabcolsep}{0.3mm}
    \renewcommand{\arraystretch}{0.8}
	\centering
	\small
	\begin{tabular}{lcccccc}		
	    \toprule		
	      & RESCAN & PReNet & DRDNet & DCSFN & RCDNet & SPDNet\\
		\midrule
		  NIQE$\downarrow$ & 3.7774 & 3.5891 & 3.8719 & 3.5326 & 3.5567 & 3.4603\\
		  PI$\downarrow$  & 2.8069 & 2.7045 & 2.8980 & 2.6427 & 2.6946 & 2.6254\\
        \bottomrule
	\end{tabular}
	\caption{Performance comparison on real-world dataset. $\downarrow$ means the better methods should achieve lower score.}
	\label{table:real}
	\vspace{-10pt}
\end{table}

We compare our proposed method  with the state-of-the-art single image deraining methods. Following RCDNet~\cite{wang2020model}, we compute PSNR and SSIM in YCbCr space. Quantitative results are shown in Table~\ref{Tab:compare} and the processing time of each method is an average time, which is calculated on 100 images of $128\times 128$ size.  It can be seen that SPDNet achieves remarkable improvements over these state-of-the-art methods. This substantiates the flexibility and generality of our proposed method in diverse rain types contained in these datasets. Furthermore, from Table~\ref{Tab:compare}, SPDNet is not only superior in performance but also superior to RCDNet~\cite{wang2020model}  and DCSFN~\cite{wang2020dcsfn} in terms of parameter and processing time.

Fig.~\ref{fig:34} illustrates the deraining performance of all competing methods on synthetic datasets. As shown, the deraining result of our proposed is better than that of other methods in sufficiently removing the rain streaks and finely recovering the image textures. For other comparison methods, they tend to blur the image textures, or still leave some visible rain streaks. 

\vspace{-12pt}

\subsubsection{Real-world Images}

In order to demonstrate that our method is also applicable in real-world scenarios, we compare with other methods on SPA-Data and a real-world dataset. Table ~\ref{Tab:compare} compare the results on SPA-Data of all competing methods quantitatively. As shown, it is easy to see that SPDNet achieves an evident superior performance than other methods.  
Moreover, since the real-world dataset supplied by Zhang~\etal~\cite{zhang2019image} does not have the corresponding label, to validate the deraining performance, we use two non-reference indicators(NIQE~\cite{mittal2012making} and PI~\cite{blau20182018}) to test the performance in this dataset. The results are shown in Table~\ref{table:real}.
It is obvious that SPDNet achieves the best results, which shows that SPDNet is not only suitable for synthetic datasets but also can repair rain in real-world scenarios. The visual results can be seen in Fig.~\ref{fig:37}. The deraining result of SPDNet is better than other methods, which finely recover the image textures and sufficiently removing the rain streaks. 

\begin{table}
    \setlength{\tabcolsep}{1.0mm}
	\renewcommand{\arraystretch}{0.8}
	\centering
	\begin{tabular}{lccccc}
	    \toprule
	   \multirow{2}{*}{ Method} & \multicolumn{2}{c}{Fusion method} & \multirow{2}{*}{Ensemble} & \multicolumn{2}{c}{Rain200H}\\
	    \cmidrule(lr){2-3} \cmidrule(lr){5-6}
		& IFM & Concat &  & PSNR & SSIM \\
		\midrule
		 w/o IFM & & $\checkmark$   & $\checkmark$ &31.05 & 0.9167\\
		 w/o Ensemble& $\checkmark$   &  &  & 30.98 & 0.9142\\
		SPDNet& $\checkmark$   &  & $\checkmark$  & 31.30 & 0.9217\\
		\bottomrule
	\end{tabular}
	\caption{Ablation study on different settings of SPDNet on Rain200H.}
	\label{table:component}
	\vspace{-12pt}
\end{table}

\subsection{Ablation Study}

\textbf{Effectiveness of the number of WMLM:}  To validate the influence of the number of WMLM on the performance, we conduct ablation experiments, which are shown in Table ~\ref{table:wmlm}. When the number of WMLM is equal to 3, the PSNR/SSIM is the best, but it will take a longer time to process the image. When the number of WMLM is equal to 1, the performance is the lowest, but compared with some current methods, it is still better and the time consumed is shorter, which shows that WMLM has a strong learning ability and may be able to serve real-time image rain removal task in future.

\begin{table}
    \setlength{\tabcolsep}{2.7mm}
    \renewcommand{\arraystretch}{0.8}
	\centering
	\begin{tabular}{ccccc}
	    \toprule
	    Numbers & PSNR & SSIM & Param & Time\\
		\midrule
		  1 & 29.85 & 0.9006 & 0.98M & 0.017s \\
		  2 & 30.90 & 0.9166 & 2.03M & 0.039s \\
		3 & 31.30  & 0.9217 & 3.04M & 0.055s \\
        \bottomrule
	\end{tabular}
	\caption{Explore the influence of different numbers of WMLM on Rain200H. The time is the average run-time, calculated on 100 image of $128\times128$ size.}
	\label{table:wmlm}
	\vspace{-8pt}
\end{table}

\textbf{Effectiveness of Basic Components:} To prove the performance of IFM and the impact of Ensemble(Sec.~\ref{ensemble}) on performance, we use Concat operation to instead the IFM and remove the Ensemble. The results are shown in Table~\ref{table:component}. SPDNet with IFM achieves significant performance improvements over SPDNet without IFM, which demonstrates that IFM is a more effective fusion method and can benefit to reconstruct clearer rain-free images.  Furthermore, SPDNet with Ensemble outperforms SPDNet without Ensemble in PSNR and SSIM, which shows that the Ensemble operation can enrich feature information so that the network can learn more useful information.

\begin{table}[t]
    \setlength{\tabcolsep}{2.7mm}
    \renewcommand{\arraystretch}{0.8}
	\centering
	\begin{tabular}{cccc}
	    \toprule
	    Iteration  & RCP Update & PSNR & SSIM\\
		\midrule
		0 & & 30.56 & 0.9144 \\
		1 & & 30.82 & 0.9161 \\
		2 & $\checkmark$ & 31.04 & 0.9197 \\
		3 & $\checkmark$ & 31.30 & 0.9217 \\
		3 & & 31.12 & 0.9191 \\
        \bottomrule
	\end{tabular}
	\caption{Explore the influence of RCP on Rain200H. RCP Update means whether to use the iterative guidance strategy to update RCP and Iteration means the number of  RCP guidance.}
	\label{table:number_guidance}
	\vspace{-14pt}
\end{table}

\begin{figure}[t]
	\begin{center}
		\includegraphics[scale=0.17]{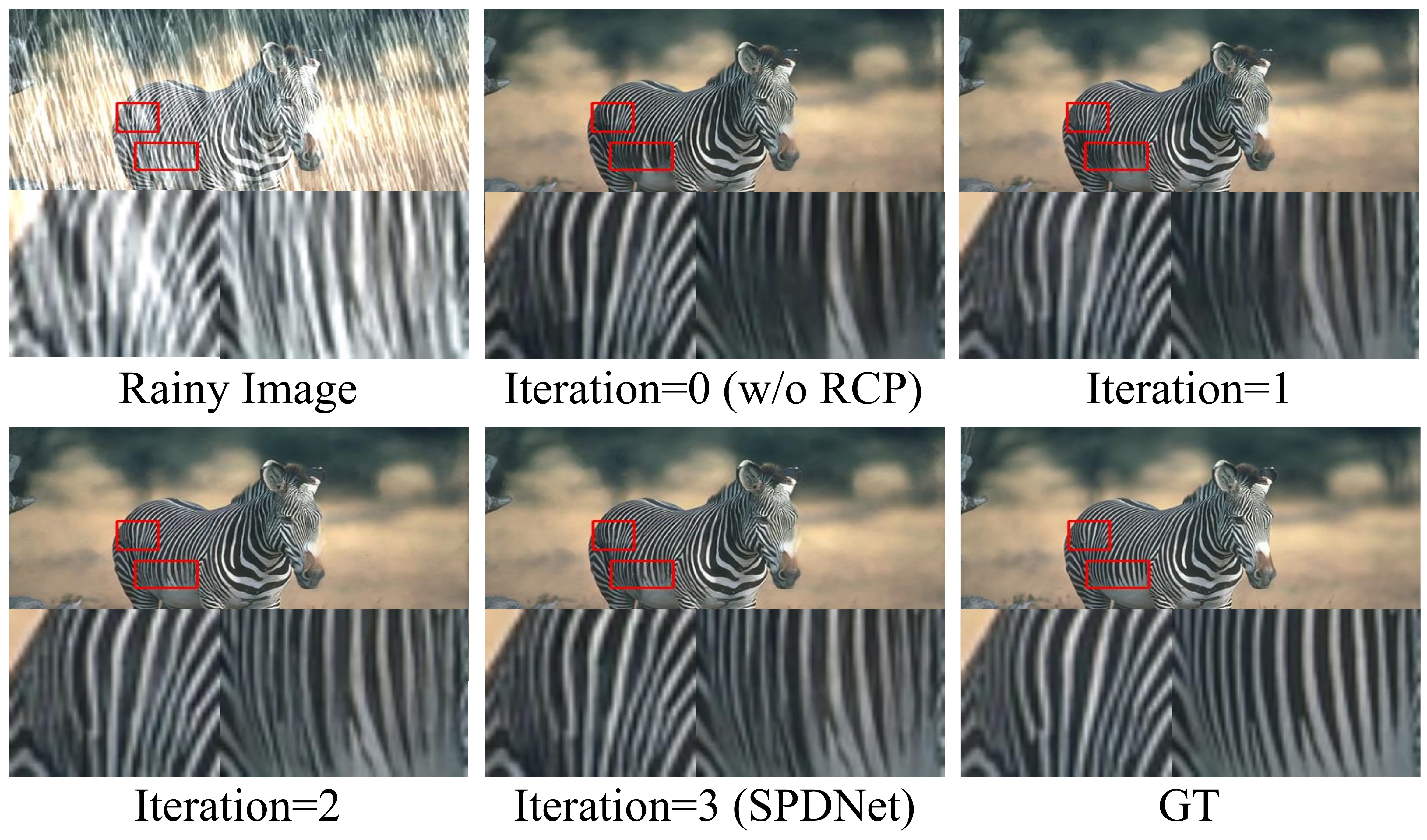}
	\end{center}
	\setlength{\abovecaptionskip}{-3pt}
	\caption{Comparison results on different numbers of RCP guidance. Obviously, compared with the method without RCP, the methods using RCR guidance can reconstruct the high-quality image with a clear and accurate structure.}
	\label{fig:guidance_number}
	\vspace{-2pt}
\end{figure}

\textbf{Effectiveness of the number of RCP guidance and Iterative Guidance Strategy:} To demonstrate that RCP can effectively assist SPDNet to generate rain-free images with clearer structure, we conduct ablation experiments and the results are shown in Table~\ref{table:number_guidance}. As shown, when more RCP guidance, the performance will be more superior. The visual comparison of different numbers of RCP guidance is shown in Fig.~\ref{fig:guidance_number}. It is observed that methods with RCP guidance can reconstruct clearer rain-free images with accurate structures, which indicates that the effectiveness of RCP in preserving structure. Moreover, to demonstrate the performance of iterative guidance strategy, we use RCP of rainy images to replace RCP of output results and the result is shown in the 4th row of Table~\ref{table:number_guidance}. It can be found that the performance of the method that only uses RCP of rainy images is lower than SPDNet, which demonstrates that  the effectiveness of iterative guidance strategy and clearer structure of RCP can reconstruct more high-quality images. \vspace{14pt}

\begin{figure*}
	\begin{center}
		\includegraphics[scale=0.5]{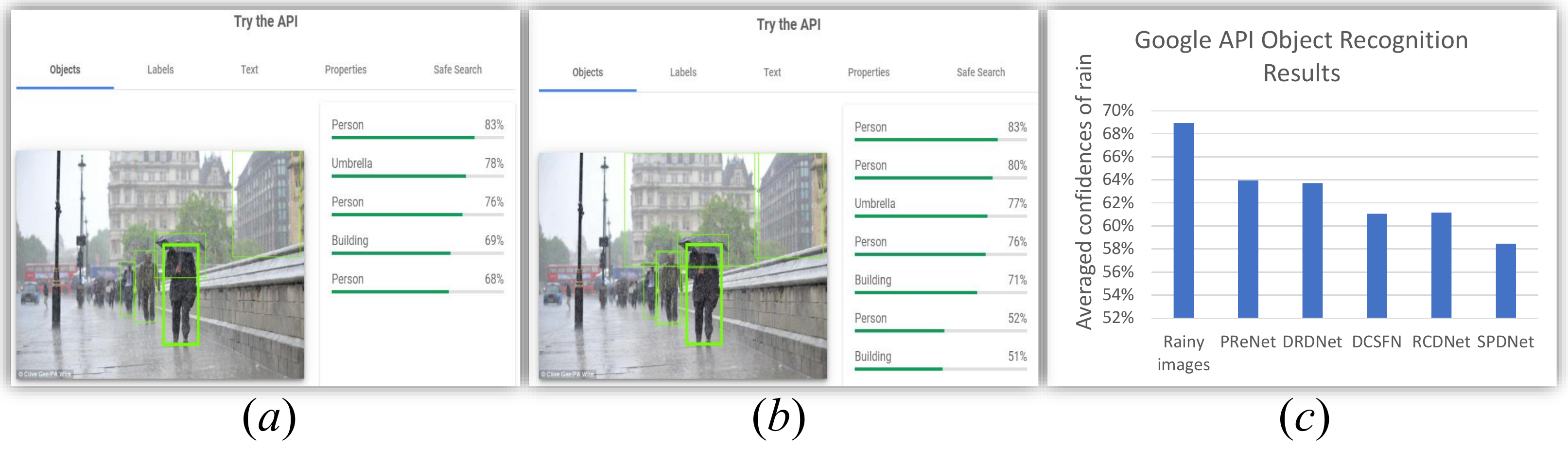}
	\end{center}
	\setlength{\abovecaptionskip}{-7.9pt}
	\caption{The deraining results are tested on the Google Vision API. \textit{a}: object recognition result in the real-world rainy image, \textit{b}: object recognition result after deraining by our proposed model, and \textit{c}: the averaged confidences in recognizing rain from 30 sets of the real-world rainy images and output results of different methods. The lower averaged confidences, the better performance of deraining.} 
	\label{fig:application}
	\vspace{-10pt}
\end{figure*}

\vspace{-13pt}

\subsection{Application}

To demonstrate that our proposed can benefit vision-based applications, we employ Google Vision API to evaluate the deraining results, as shown in Fig.~\ref{fig:application}. Fig.~\ref{fig:application} (\textit{a}) illustrates the object recognition result in the real-world rainy image and Fig.~\ref{fig:application} (\textit{b}) shows the object recognition result in the output result of SPDNet. Comparing Fig.~\ref{fig:application} (\textit{a}) and (\textit{b}), the number of detections in our result is more, indicating that SPDNet can effectively improve the performance of detection and recognition. Furthermore, according to Deng~\etal~\cite{deng2020detail}, we use Google Vision API to detect the confidences of rain in 30 sets of real-world rainy images and the results of five methods, as shown in Fig.~\ref{fig:application}\,\,(\textit{c}). The confidence of rain means that the probability of rainy weather. When confidence is lower, the rain is more light, which shows that the better deraining performance. As one can see, the confidence in recognizing rain from the output results of SPDNet is significantly reduced, which indicates that the deraining performance of SPDNet outperforms other methods in real-world datasets. Furthermover, we also use YOLOv3 to perform detection on COCO350~\cite{jiang2020multi}, as shown in Table \ref{table:coco}(mAP-50 means the mAP at .5 IOU metric). Compared with other models, the results generated by SPDNet effectively promote the detection performance, which further demonstrates the effectiveness of our proposed SPDNet.

\begin{table}
	\setlength{\tabcolsep}{1.3mm}
	\renewcommand{\arraystretch}{1}
	\centering
	\small
	\begin{tabular}{cccccc}		
		\hline		
		& Rainy & DRDNet & DCSFN & RCDNet & SPDNet\\
		\hline
		mAP-50$\uparrow$ & 0.551 & 0.652 & 0.665 & 0.658 & \textbf{0.692}\\
		\hline
	\end{tabular}
	\caption{The detection performance comparison on COCO350 dataset~\cite{jiang2020multi}. $\uparrow$ means the better methods should achieve higher score.}
	\label{table:coco}
	\vspace{-14pt}
\end{table}

\vspace{-3pt}
\section{Conclusion}
In this paper, we have proposed a Structure-Preserving Deraining Network (SPDNet) with residue channel prior guidance. To achieve this, an RCP guidance network and an iterative guidance strategy are proposed for structure-preserving deraining. Specifically, an effective WMLM is proposed as the backbone to fully learn the background information. Meanwhile, the RCP is introduced as reference information to guide the learning of WMLM, and IFM is designed to make full use of the RCP information. In addition, based on an observation that the extracted RCP also shows better results with the improvement of deraining image quality, an iterative update strategy is proposed to improve the accuracy of RCP and then re-guide the learning of WMLM. The ablation study demonstrates the performance of RCP, IFM, and iterative update strategy. Experimental results on several synthetic deraining datasets and real-world scenarios have shown the great superiority of our proposed SPDNet over other top-performing methods.

\noindent \textbf{Acknowledgment.} This work was supported by the Key Project of the National Natural Science Foundation of China under Grant 61731009, the NSFC-RGC under Grant 61961160734, the National Natural Science Foundation of China under Grant 61871185, the Shanghai Rising-Star Program under Grant 21QA1402500, and the Science Foundation of Shanghai under Grant 20ZR1416200. The work of Tieyong Zeng was supported by the National Key R\&D Program of China under Grant 2021YFE0203700, Grant NSFC/RGC N\_CUHK 415/19, Grant RGC 14300219, 14302920, 
14301121, and CUHK Direct Grant for Research under Grant 4053405, 4053460.

\newpage
{\small
\bibliographystyle{ieee_fullname}
\bibliography{egbib}
}

\end{document}